%% file: main.tex
\documentclass[10pt,twocolumn,letterpaper]{article}

\usepackage[pagenumbers]{cvpr} 

\usepackage{multirow}
\usepackage{multicol}
\usepackage{xcolor}
\usepackage{pifont}
\usepackage[table]{xcolor}
\usepackage{titletoc}
\usepackage[most]{tcolorbox} 
\usepackage{marvosym}

\definecolor{cvprblue}{rgb}{0.21,0.49,0.74}
\usepackage[pagebackref,breaklinks,colorlinks,allcolors=cvprblue]{hyperref}

\def\paperID{1285} 
\def\confName{CVPR}
\def\confYear{2026}

\definecolor{darkorange}{rgb}{1.0, 0.55, 0.0}
\definecolor{mygreen}{RGB}{0,180,0}
\definecolor{myred}{RGB}{180,0,0}  
\newcommand{\icook}{\textcolor{mygreen}{\ding{51}}} 
\newcommand{\icox}{\textcolor{myred}{\ding{55}}}    
\newcommand{\icohalf}{\textcolor{darkorange}{\ding{51}\kern-0.65em\ding{55}}}
\newcommand{\benchmark}{RULER-Bench\xspace}

\newtcolorbox[auto counter, number within=section]{promptbox}[2][]{%
  colback=white,
  colframe=green!50!gray!40!black,
  width=\textwidth,
  arc=1mm,
  boxrule=0.5mm,
  title={\normalsize#2},
  #1
}

\title{\benchmark: Probing Rule-based Reasoning Abilities of Next-level Video Generation Models for Vision Foundation Intelligence}

\author{
  \bf{Xuming He}\textsuperscript{1*} \quad 
  \bf{Zehao Fan}\textsuperscript{1*} \quad 
  \bf{Hengjia Li}\textsuperscript{1*\dag}  \quad 
  \bf{Fan Zhuo}\textsuperscript{1} \quad 
  \bf{Hankun Xu}\textsuperscript{1} \\
  \bf{Senlin Cheng}\textsuperscript{2}  \quad
  \bf{Di Weng}\textsuperscript{1} \quad
  \bf{Haifeng Liu}\textsuperscript{1} \quad
  \bf{Can Ye}\textsuperscript{2} \quad
  \bf{Boxi Wu}\textsuperscript{1} \\
\textsuperscript{1}Zhejiang University  \quad
\textsuperscript{2}Ant Group  \\
\\
[-10pt]
\tt\normalsize\color{Blue}\url{https://hexmseeu.github.io/RULER-Bench-proj/}
}

\begin{document}
\twocolumn[{%
\renewcommand\twocolumn[1][]{#1}%
\maketitle

\maketitle
\vspace{-2em}
\includegraphics[width=\linewidth]{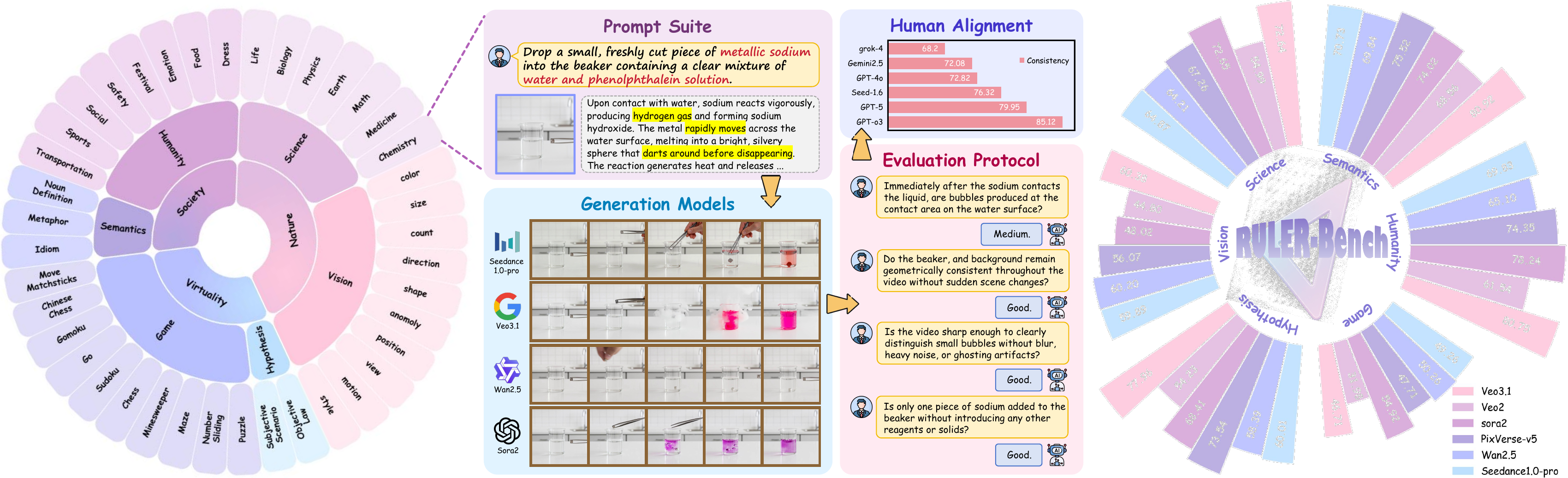}
\captionof{figure}{\textbf{Overview of \benchmark.} 
    We propose \benchmark, a comprehensive benchmark designed to evaluate the rule-based reasoning abilities of video generation models.
    \textbf{\textit{Left:}} Grounded in three fundamental domains, we formulate rule-based reasoning ability into six categories: \textit{Science}, \textit{Vision}, \textit{Hypothesis}, \textit{Game}, \textit{Semantics}, and \textit{Humanity}.
    These categories are further subdivided into 40 tasks.
    \textbf{\textit{Center:}} Using the collected samples, we evaluate 10 video models based on the corresponding checklist across four metrics. 
    Each checklist question is scored by GPT-o3 with discrete labels.
    To validate the reliability of the evaluator, we conduct a human alignment study, in which GPT-o3 achieves 85\% agreement with human judgments.
    \textbf{\textit{Right:}} Extensive experiments demonstrate that Veo3.1 achieves the best performance.
    However, all models exhibit limited reasoning ability across different rule categories.
    \vspace{1.5em}
}
\label{fig:teaser}
}]

\input{sec/00_abstract}    
\input{sec/01_introduction}
\input{sec/02_related_work}
\input{sec/03_data}
\input{sec/04_eval}

\input{sec/05_exp}
\input{sec/06_conclusion}

{
    \small
    \bibliographystyle{ieeenat_fullname}
    \bibliography{main}
}
\input{sec/07_supp}


\end{document}

%% file: sec/00_abstract.tex
\begin{abstract}
Recent advances in video generation have enabled the synthesis of videos with strong temporal consistency and impressive visual quality, marking a crucial step toward vision foundation models.
To evaluate these video generation models, existing benchmarks primarily focus on factors related to visual perception and understanding, like visual aesthetics, instruction adherence, and temporal coherence. 
However, the rule-based reasoning capabilities of video generation models remain largely unexplored. 
Although recent studies have carried out preliminary explorations into whether video models can serve as zero-shot learners, they still lack a fine-grained decomposition of reasoning capabilities and a comprehensive evaluation protocol.
%
\makeatletter{\renewcommand*{\@makefnmark}{}
\footnotetext{\textsuperscript{*}equal contributions. \quad
\textsuperscript{\dag}project lead.
%
}
}
To address this gap, we introduce \benchmark, a benchmark designed to evaluate the reasoning ability of video generation models from the perspective of cognitive rules.
Built upon two fundamental paradigms: text-to-video and image-to-video, \benchmark covers 40 representative tasks spanning six rule categories with 622 high-quality annotated instances.
For the evaluation of each generated video, we construct a checklist covering four metrics and leverage GPT-o3 to assign scores to each question, achieving 85\% alignment with human judgements.
Extensive experiments show that the state-of-the-art model achieves only 48.87\% on the rule coherence metric, highlighting significant room for improvement in the reasoning capability of next-level video models. 
We expect that the insight obtained from \benchmark will facilitate further development of reasoning-aware video generation, advancing video generation models toward vision foundation intelligence.

\end{abstract}

%% file: sec/01_introduction.tex
\vspace{-1.5em}
\section{Introduction}
\label{sec:intro}

Video generation focuses on producing video clips that exhibit strong temporal consistency and high visual quality, serving as a fundamental technique for downstream applications such as customization~\cite{li2025magicid,wei2025dreamrelation,li2025personalvideo,ye2025stylemaster} and world modeling~\cite{agarwal2025cosmos,wang2025internvideo2,duan2025worldscore}.
Empowered by advances in generative frameworks such as diffusion models~\cite{peebles2023scalable,ho2020denoising,zhang2023adding,rombach2022high,gao2024lumina,blattmann2023align} and autoregressive~\cite{tian2024visual,jin2024pyramidal,teng2025magi} approaches, recent video generation systems have achieved remarkable progress on perceptual and understanding abilities, such as aesthetic quality and instruction adherence.
%
%
However, with the rapid scaling of high-quality datasets and model parameters, state-of-the-art models such as Sora2~\cite{openai2025sora2}, Veo3~\cite{GoogleDeepMind2025Veo3}, and Wan2.5~\cite{wan2.5-model2025} have nearly saturated on these dimensions.

%
Mirroring the evolution of foundation language models~\cite{bai2025qwen2,yang2024qwen2llm,bi2024deepseek,glm2024chatglm,touvron2023llama,guo2025deepseek} in natural language processing, video models are expected to gradually advance from perception and understanding to reasoning, thereby paving the way for ultimately becoming the foundation model for vision.
\cref{fig:teaser} illustrates the current stage of this evolution. Given the instruction \textit{``drop a piece of sodium into the phenolphthalein solution''} and an input image, all four generation models produce visually coherent video clips. 
However, their reasoning abilities differ significantly.
For instance, Veo3.1~\cite{GoogleDeepMind2025Veo3} accurately infers that sodium reacts violently with phenolphthalein solution, releasing gas and forming an alkaline product. 
Therefore, the generated video depicts vigorous bubbling and the distinctive red coloration.
In contrast, Seedance1.0-pro~\cite{gao2025seedance}, Wan2.5, and Sora2 exhibit limited reasoning ability, capturing only a subset of the expected reaction phenomena.
This capability gap reveals a fundamental challenge: \textit{How to assess whether next-level video generation models possess the reasoning abilities necessary to achieve vision foundation intelligence?}

Recently, Wiedemer et al.~\cite{wiedemer2025video} have conducted an initial exploration into whether Veo3 could serve as a vision foundation model.
They constructed instances spanning four aspects: perception, modeling, manipulation, and reasoning, and then measured Veo3's success rates on each dimension.
However, their exploration lacks a fine-grained decomposition of reasoning characteristics and a systematic evaluation framework. 
Additionally, VBench-2.0~\cite{zheng2025vbench} and PhyGen-bench~\cite{meng2024towards} target intrinsic faithfulness and physical laws, but remain limited to physics and commonsense reasoning, overlooking the diversity of reasoning dimensions.

To bridge this gap, we conceptualize reasoning as \textbf{cognitive rule-based prediction}, the ability of a generative model to infer rules from inputs and apply them to predict plausible phenomena in videos.
Building on this formulation, we organize reasoning scenarios into three fundamental domains: Nature, Society, and Virtuality, which correspond to real-world, human-centered, and virtual environments, respectively.
Within these domains, we further define six categories of cognitive rules: Vision, Science, Semantics, Hypothesis, Game, and Humanity, which collectively span diverse reasoning scenarios in video generation.

Based on these categories, we propose \benchmark (\textbf{RULE}-based \textbf{R}easoning \textbf{Bench}mark for video generation), a comprehensive benchmark designed to evaluate the reasoning capabilities of next-level video generation models in two typical generation scenarios: text-to-video and image-to-video.
\benchmark adopts a hierarchical paradigm, organizing 40 tasks within six rule categories.
To ensure evaluation reliability, \benchmark comprises 622 high-quality instances, evenly distributed across tasks.

Furthermore, leveraging multimodal large language models (MLLMs), we introduce a comprehensive evaluation protocol that assesses generated videos across four metrics: Instruction Following, Visual Consistency, Visual Fidelity, and Rule Coherence.
Unlike subjective continuous scoring, we construct a checklist of objective questions for each instance based on these four metrics.
Each question is then rated on a discrete three-point scale: \textit{good}, \textit{medium}, or \textit{bad}.
To validate the reliability of our evaluation protocol, we manually annotate 813 checklist questions based on generated videos and verify the consistency between MLLM responses and human judgments.
As shown in~\cref{fig:teaser}, our evaluation protocol achieves alignment with human annotation on 85\% of the checklist questions, which confirms its effectiveness.
Building upon \benchmark, we further conduct a systematic evaluation of 10 state-of-the-art video generation models. 
Extensive experiments show that all models consistently exhibit limitations in rule coherence, demonstrating substantial potential for enhancing the reasoning capability of next-level video models.

The main contributions of this work are:

\begin{itemize}
    \item We conceptualize reasoning in video generation as \textit{cognitive rule-based prediction} and formulate the first taxonomy that organizes reasoning into six rule categories.
    \item We introduce \benchmark, a comprehensive benchmark specifically designed to evaluate the reasoning abilities of video generation models, and systematically covering 40 tasks and 622 high-quality instances.
    \item We conduct extensive experiments on 10 state-of-the-art video generation models, revealing substantial limitations across all models across different rule types.
\end{itemize}

%% file: sec/02_related_work.tex
\section{Related Work}
\label{sec:rel}

\textbf{Video Generation Models}. Recent advancements in diffusion models~\cite{esser2024scaling,ho2020denoising,ho2022imagen,ho2022video,nichol2021improved,song2020denoising,song2020score,zhang2023adding,zhang2025tcsinger,zhang2025versatile} and autoregressive approaches~\cite{hong2022cogvideo,tian2024visual,li2024controlvar,jin2024pyramidal,xie2024show,chen2025janus} have led to rapid progress in video generation~\cite{guo2023animatediff,blattmann2023align,wang2025lavie,liu2025improving,henschel2025streamingt2v,wu2025customcrafter,lin2024open,gupta2024photorealistic,ma2024latte,jiang2024videobooth}. 
By leveraging large-scale, high-quality training data and expanding model capacity, recent systems such as Sora2, Veo3, and Wan have achieved remarkable performance in the dimensions of perception and understanding, including visual consistency, aesthetic quality, and instruction adherence.
As visual fidelity continues to improve, research attention has begun to shift focus towards exploring reasoning capabilities~\cite{agarwal2025cosmos,zhang2025videorepa,wang2025wisa,chen2025towards,xue2025phyt2v,zhang2024physdreamer,liu2024physgen,gao2024flip,xie2025physanimator,montanaro2024motioncraft}, such as understanding physical laws and performing logical inference.
NewtonGen~\cite{yuan2025newtongen} introduces Neural Newtonian Dynamics to model and predict Newtonian motions, while V-Chain~\cite{huang2025vchain} incorporates a chain-of-visual-thought mechanism to inject visual reasoning signals into generative processes.
However, existing benchmarks lack a systematic framework to assess the reasoning abilities of video models. To address this challenge, \benchmark provides a comprehensive benchmark for evaluating the emerging capabilities of video generation models for rule-based reasoning.

\begin{figure*}[t]
  \centerline{\includegraphics[width=\textwidth]{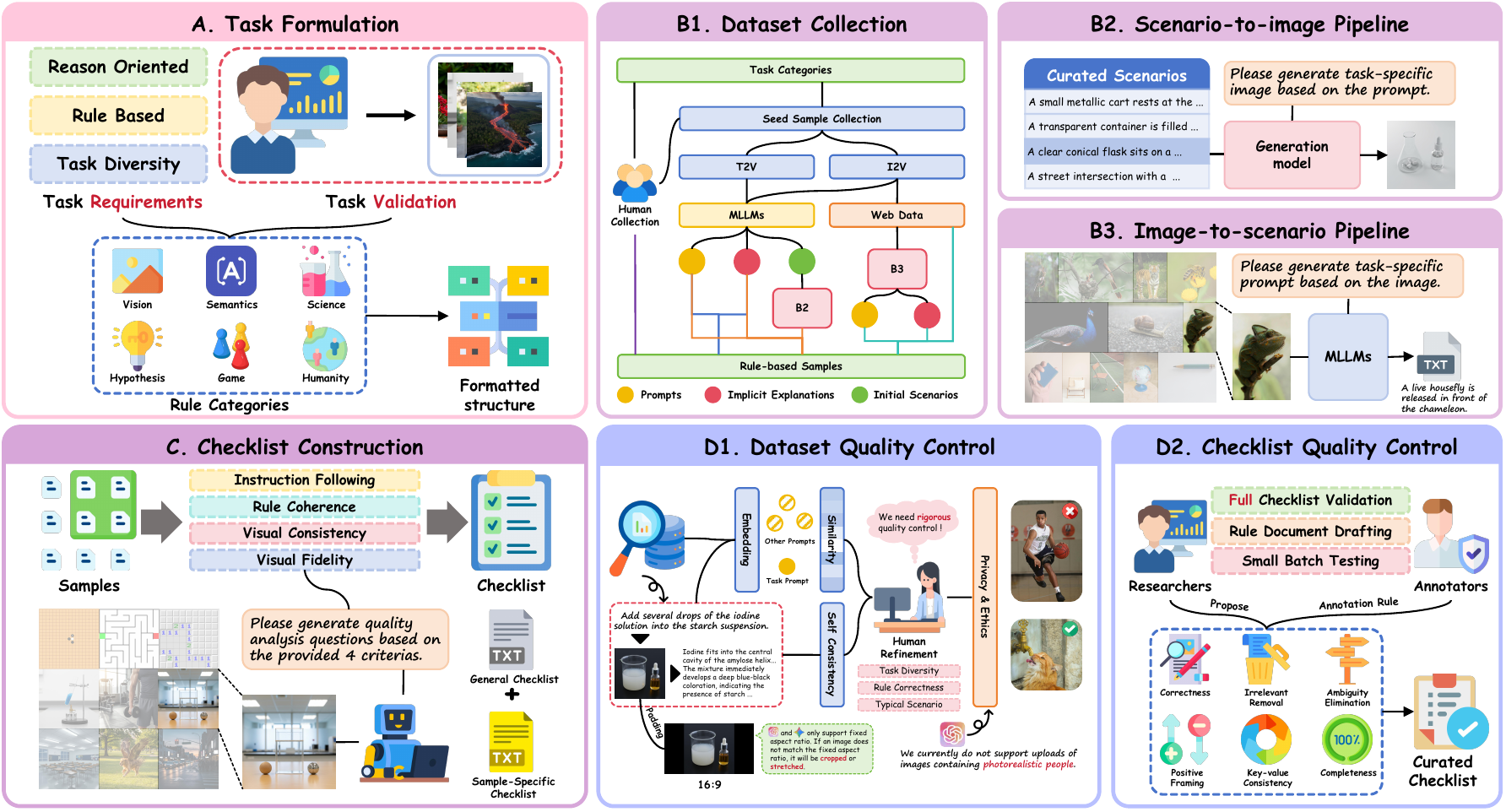}} 
  \caption{\textbf{Overview of dataset construction and validation}.
  First, we formulate our tasks based on the six rule categories.
  Second, we design task-specific data construction pipelines for T2V and I2V tasks.
  Third, we leverage MLLM to construct checklist questions across four evaluation metrics.
  Finally, we conduct quality control and data refinement for the constructed dataset and checklists.
  }
  \label{fig:data}
\end{figure*}

\textbf{Benchmarks for Video Generation Methods}. To effectively evaluate the capabilities of video generation models, a variety of benchmarks have been proposed.
VBench~\cite{huang2024vbench}, EvalCrafter~\cite{liu2024evalcrafter}, and FetV~\cite{liu2023fetv}  primarily focus on Text-to-Video (T2V) tasks, assessing fundamental technical attributes.
In contrast, T2V-CompBench~\cite{sun2025t2v}, PhyGen-bench~\cite{meng2024towards}, and VBench-2.0~\cite{zheng2025vbench} focus on deeper principles such as compositionality, physical law, and intrinsic faithfulness.
For Image-to-Video (I2V) tasks, benchmarks such as AIGC-Bench~\cite{fan2023aigcbench} and AnimateBench~\cite{fan2023aigcbench} evaluate the video generation models along perceptual dimensions, including instruction following and visual consistency, while UI2V-Bench~\cite{zhang2025ui2v} evaluates models from a general understanding perspective.
Regarding Video-to-Video (V2V) generation, VE-Bench~\cite{sun2025ve} and EditBoard~\cite{chen2025editboard} cover general editing scenarios, including subject, style, and attribute editing, whereas IVE-Bench~\cite{chen2025ivebench} integrates traditional metrics with large language model-based assessments across multiple editing categories.
However, as shown in~\cref{tab:related_work}, these benchmarks are typically constrained to a single task type and primarily concentrate on fundamental abilities or specific emerging capabilities. 
Systematic evaluation of reasoning dimensions remains largely unexplored, with insufficient evaluation data~\cite{guo2025video} and a lack of formal rule-based frameworks~\cite{wiedemer2025video}. 
\benchmark unifies two task paradigms: T2V and I2V, and provides a comprehensive evaluation of rule-based reasoning in video models, addressing limitations in existing evaluation practices and advancing the field toward vision foundation models.

%% file: sec/03_data.tex
\section{\benchmark}
\label{sec:data}

\subsection{Rule-Based Task Formulation}
Inspired by KRIS-Bench~\cite{wu2025kris}, a benchmark that evaluates image editing models through the lens of knowledge, we conceptualize reasoning in video generation as \textbf{cognitive rule-based prediction}.
Specifically, video models are required to infer implicit rules from the given input, predict possible outcomes, and present them through the generated video.
Grounded in the three fundamental domains of the world: Nature, Society, and Virtuality, we define six rule categories that collectively characterize diverse and complex reasoning scenarios in video generation.


\textbf{Natural Domain} requires video generation models to perform reasoning grounded in real-world rules, thereby generating videos with plausible visual phenomena.
The rule categories in this domain include:
\begin{itemize}
    \item \textit{Vision}. The ability of video models to reason about real-world visual compositions, such as appearance variations, spatial arrangements, and dynamic transformations.
    \item \textit{Science}. The ability of models to reason about scientific phenomena by leveraging the underlying laws of nature.
\end{itemize}

\textbf{Social Domain} challenges video models to reason grounded in rules derived from human society, generating videos exhibiting socially consistent behaviors and interactions. The rule categories in this domain encompass:
\begin{itemize}
    \item \textit{Humanity}. The ability of video models to infer human behaviors by leveraging principles of social dynamics, such as sports rules, traditional customs, and festivals.
    \item \textit{Semantics}. The ability of video models to reason about authentic expressions through semantic regularities.
\end{itemize}

\textbf{Virtual Domain} requires video models to perform reasoning grounded in rules of virtual environments, generating videos that follow the internal logic of these environments. The rule categories within this domain include:
\begin{itemize}
    \item \textit{Game}. The ability of video models to perform rational actions according to game rules to win matches or complete levels, for example, executing a checkmate in chess.
    \item \textit{Hypothesis}. The ability of video models to reason about and predict phenomena based on hypothetical premises.
\end{itemize}

For each rule category, we carefully design distinct tasks to enable a finer-grained decomposition. 
As shown in~\cref{fig:teaser}, \benchmark encompasses a total of 40 tasks, with detailed descriptions provided in the Appendix.

\begin{table}
  \caption{Comparison of open-source video generation benchmarks. \icohalf~represents insufficient reasoning dimensions.}
  \label{tab:related_work}
  \centering
  \footnotesize
  \setlength\tabcolsep{4.1pt}
\begin{tabular}{@{}lcccccc@{}}
\toprule
\textbf{Benchmark}     & \textbf{Size} & \textbf{Categories} & \textbf{Tasks} & \textbf{Type} & \textbf{Reason} & \textbf{Rule} \\ \midrule
FetV~\cite{liu2023fetv}    & 619 & 3          & 22   & T2V      & \icox          & \icox           \\
EvalCrafter~\cite{liu2024evalcrafter}    & 700 & 4          & 12   & T2V      & \icox          & \icox           \\
TC-Bench~\cite{feng2024tc} & 270 & -          & 3    & T/I2V     & \icox          & \icox           \\
VBench-2.0~\cite{zheng2025vbench}    & 1260 & 5          & 18    & T2V      & \icohalf          & \icox           \\
VideoPhy~\cite{bansal2024videophy}    & 688 & 3          & 6   & T2V      & \icohalf          & \icox           \\
UI2V-Bench~\cite{zhang2025ui2v}    & 500  & 4          & 21    & I2V      & \icohalf          & \icox           \\
MME-CoF~\cite{guo2025video}       & 59   & -          & 12    & T/I2V    & \icook          & \icox           \\ \midrule
\textbf{\benchmark}   & 622  & 6          & 40    & T/I2V  & \icook          & \icook            \\ \bottomrule
\end{tabular}
\vspace{-15pt}
\end{table}

\subsection{Data Construction}
\benchmark encompasses two task paradigms: Text-to-Video (T2V) and Image-to-Video (I2V).
To construct a benchmark with high-quality and diverse coverage, we curate the data through a hybrid process that integrates human annotation with GPT-5~\cite{openai2025gpt5} generation.
%
%
In this section, we provide a detailed elaboration on the data construction procedures for each task paradigm. An overview of the data construction pipeline is illustrated in Part B of~\cref{fig:data}.

\textbf{Seed sample collection}. We first manually curate seed samples for each task to provide GPT-5 with in-context guidance,  which enables it to better comprehend task requirements and thereby generate more reliable samples.

\textbf{T2V data collection}. For each T2V instance, we first define a tuple (\textit{prompt, implicit explanation}), where the prompt serves as the sole input to the video generation models, and the implicit explanation specifies the expected outcome along with the underlying reasoning principles. 
Second, we provide GPT-5 with detailed task requirements and a set of seed samples, and instruct it to generate new samples that strictly adhere to the defined criteria, thereby ensuring both diversity and consistency in the dataset.

\textbf{I2V data collection}. Traditional image collection for generative model benchmarks relies on web-sourced images~\cite{wu2025kris,zi2025se}. 
However, these images are largely confined to generic scenes, such as animals or cultural landscapes, while providing minimal coverage of domain-specific content.
With the recent advances in image generation, state-of-the-art models~\cite{janus,labs2025flux,deng2025emerging} can generate richly detailed images based on highly customizable prompts.
Therefore, we collect I2V data by integrating web-sourced and generated images based on the characteristics of each task.

For instances containing web images, we adopt the same procedure as in T2V by defining a tuple (\textit{prompt, implicit explanation}).
For each image, GPT-5 is used to design the prompt and implicit explanation based on the image content and task requirements.
However, for samples in the Game category, MLLMs often struggle to generate implicit explanations that correctly reflect the underlying game strategies.
To address this limitation, we collect \textit{ground truth images} representing the final frame of the expected outcome, and manually create the corresponding implicit explanations to accurately capture the underlying rules. 

For samples containing generated images, we define a triplet (\textit{initial scenario, prompt, implicit explanation}), with the initial scenario serving as the input to the text-to-image (T2I) model.
To ensure consistency, we instruct GPT-5 to leverage the initial scenario as the starting conditions when generating the prompt.
Additionally, to maintain high visual quality, GPT-5 is guided to design initial scenarios that prevent the T2I model from generating images containing textual content, which remains challenging in current T2I systems.
We employ the powerful nano banana~\cite{nano-banana-2025} for image generation, providing both the generated image and the corresponding prompt as inputs to the I2V tasks.

\subsection{Quality Control and Data Refinement}
To ensure high data quality, we design a meticulous quality control pipeline, illustrated in Part D1 of~\cref{fig:data}. 
First, we use a robust embedding model to encode all prompts into text embeddings.
Second, we compute cosine similarity between prompts in the feature space. 
Highly similar prompts are deduplicated.
To further ensure content fidelity, we instruct MLLM to verify consistency among the prompt, implicit explanation, and the corresponding input image.
We then perform human refinement on instances that are deduplicated and self-consistent, with each sample carefully validated for rule correctness, task diversity, and scenario typicality.
Finally, we pre-generate the verified samples using Sora2 to filter out instances subject to privacy or ethical constraints.
During generation, we observe that Sora2 generates videos with fixed aspect ratios (16:9 or 9:16), which may cause stretching or cropping.
To prevent information loss, we pad each input to match the target aspect ratio while preserving all critical visual content.

%% file: sec/04_eval.tex
\section{Evaluation Protocol}
\label{sec:eval}

\subsection{Evaluation Metrics}
To comprehensively evaluate the performance of state-of-the-art video models on \benchmark, we consider 3 commonly used metrics: \textit{Instruction Following}, \textit{Visual Consistency}, and \textit{Visual Fidelity}~\cite{guo2025video,zhao2025envisioning,wu2025kris}.
Additionally, we introduce a novel rule-based dimension, termed \textit{Rule Coherence}, which assesses whether the generated video adheres to the implicit rules underlying the given prompt.
The detailed definitions of the 4 metrics are presented below.

\textbf{Instruction Following} evaluates whether video models generate videos that faithfully follow the user's instructions. 
It focuses on the semantic alignment with the input prompt and is independent of implicit rules.
For instance, given the prompt ``place a small wooden sphere and an iron sphere into water'', the Instruction Following metric examines whether the balls are placed into water, without considering the subsequent phenomena or physical principles.

\textbf{Visual Consistency} evaluates whether video generation models preserve identities of elements that are expected to remain unchanged throughout the generated video.
For example, in sports scenes, the Visual Consistency dimension focuses on whether visual attributes, such as stadium color, remain consistent across the entire video.

\textbf{Visual Fidelity} assesses the overall visual quality of the generated video, including whether it is visually clear, stable, and free from noise, artifacts, or distortions.

\textbf{Rule Coherence} assesses whether the generated video adheres to scene-specific rules.
It requires video models to leverage implicit rules to predict or infer visual phenomena from the given instructions.
For example, given the prompt ``add extra water to the left arm of a transparent U-shaped tube and observe the water levels in both arms'', generation models are expected to apply the principle of communicating vessels to infer that the final water levels remain equal and depict this phenomenon accurately in the video.

\begin{figure}[t]
  \centering
  \includegraphics[width=\linewidth]{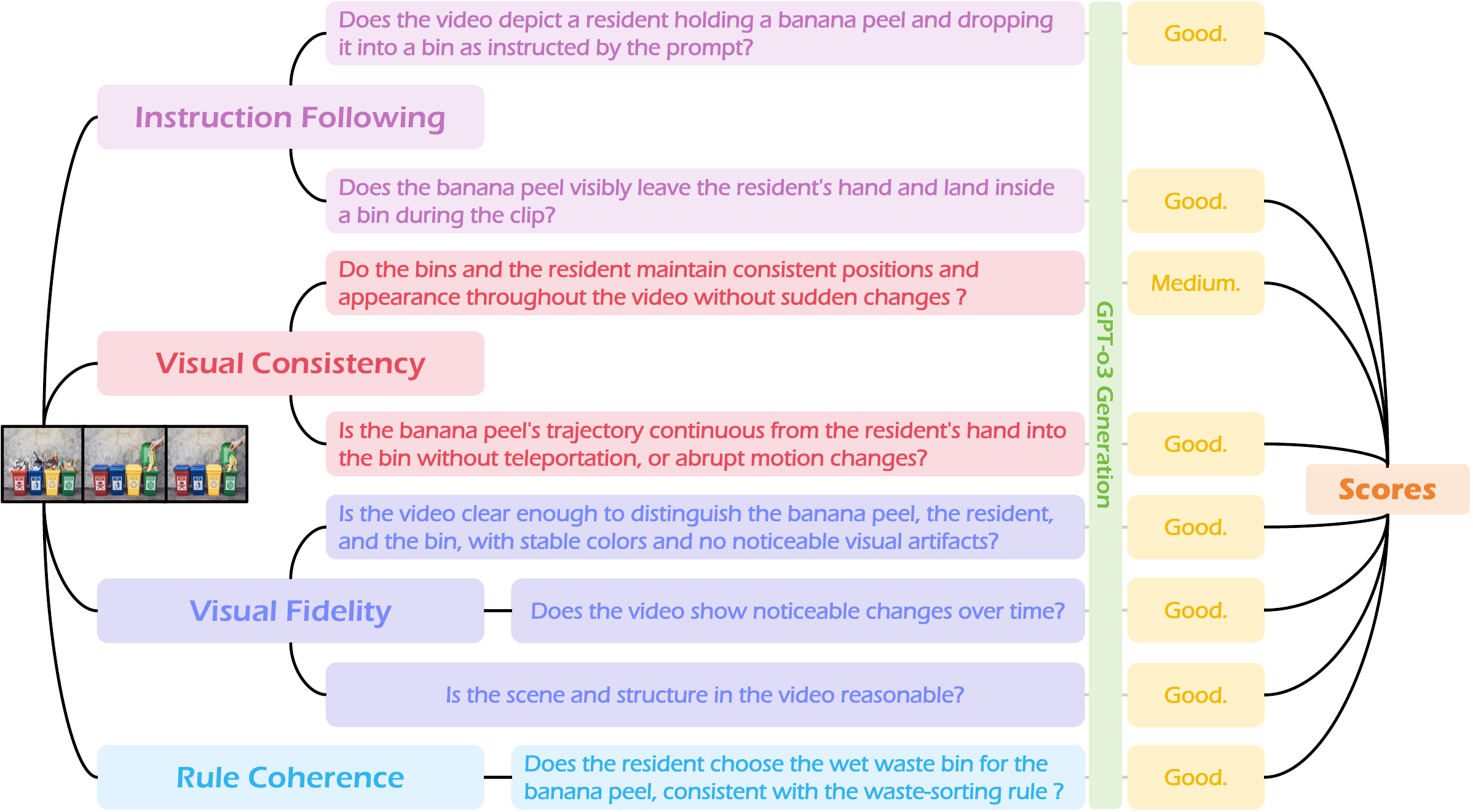}
  \caption{Evaluation pipeline of \benchmark.
  }
  \label{fig:eval}
  \vspace{-15pt}
\end{figure}

Inspired by~\cite{li2025easier}, we design a checklist-based evaluation for each sample.
As shown in~\cref{fig:eval}, each checklist contains multiple questions derived from the four evaluation metrics, and each question is answered using one of the three options: \textit{good}, \textit{medium}, or \textit{bad}.
Compared to traditional scoring methods~\cite{wu2025kris,sun2025t2v}, the checklist protocol provides greater interpretability, allowing evaluators to make more consistent and human-aligned judgements.
Given the recent advances in MLLMs of video understanding, we employ o3~\cite{openai2025o3} as the evaluator to assess the rule-based reasoning ability of video generation models.

\subsection{Checklist Construction and Quality Control}
As shown in~\cref{fig:data}, we leverage MLLMs to generate evaluation checklists.
For each sample, we instruct GPT-5 to generate multiple questions based on the input image, prompt, and implicit explanation, covering the four evaluation metrics.
Each question is designed to evaluate the generated video from a distinct perspective, ensuring that diverse aspects of video content are systematically assessed.
Detailed instruction prompts are provided in the Appendix.

\begin{table*}[ht]
\centering
\setlength\tabcolsep{6.3pt}
\footnotesize
\caption{
    \textbf{Evaluation result across different rule categories and metrics}, including Instruction Following (IF), Visual Consistency (VC), Visual Fidelity (VF), and Rule Coherence (RC). 
    All models exhibit limited rule-based reasoning ability.
    The performance of closed-source models and open-source models is separately marked with the best results in \textbf{bold}, and the second best \underline{underlined}.
}
\label{tab:main_result}
\begin{tabular}{cc|cccccc|cccc}
\toprule
\multicolumn{1}{c|}{\multirow{2}{*}{\textbf{Rule Categories}}} & \multirow{2}{*}{\textbf{Metric}} & \multicolumn{6}{c|}{\textbf{Closed-Source Models}}                                                                                                                                              & \multicolumn{4}{c}{\textbf{Open-Source Models}}                                                                                                                                                                                            \\ \cmidrule(l){3-12} 
\multicolumn{1}{c|}{}                                 &                         & Veo3.1        & Veo2       & Sora2         & \begin{tabular}[c]{@{}c@{}}Pixel\\ Verse-V5\end{tabular} & Wan 2.5      & \begin{tabular}[c]{@{}c@{}}Seedance\\ 1.0-pro\end{tabular} & \begin{tabular}[c]{@{}c@{}}Hunyuan\\ Video\end{tabular} & \begin{tabular}[c]{@{}c@{}}CogVideoX\\ 1.5 5B\end{tabular} & \begin{tabular}[c]{@{}c@{}}Wan 2.2\\ A14B\end{tabular} & \begin{tabular}[c]{@{}c@{}}Wan 2.1\\ 14B  \end{tabular} \\ \midrule
\multicolumn{1}{c|}{\multirow{5}{*}{Science Rule}}    & IF                      & {\underline{65.05}}    & 42.17       & \textbf{66.00} & 57.13                                                    & 57.38          & 58.86                                                      & 24.92                                                   & 27.97                                                   & \textbf{37.15}                                                 & \underline{35.80}                                                 \\
\multicolumn{1}{c|}{}                                 & VC                      & {\underline{83.18}}    & 73.3        & \textbf{88.01} & 80.76                                                    & 80.48          & 80.99                                                      & 48.46                                                   & 48.84                                                   & \textbf{68.52}                                                 & \underline{65.74}                                                 \\
\multicolumn{1}{c|}{}                                 & VF                      & \textbf{91.37} & 82.33       & 89.49          & {\underline{89.74}}                                              & 85.35          & 87.69                                                      & 71.29                                                   & 70.96                                                   & \underline{80.37}                                                 & \textbf{81.93}                                                 \\
\multicolumn{1}{c|}{}                                 & RC                      & \textbf{50.97} & 22.16       & {\underline{47.09}}    & 41.41                                                    & 33.64          & 31.96                                                      & 12.64                                                   & 13.70                                                   & \textbf{17.16}                                                 & \underline{15.90}                                                 \\
\multicolumn{1}{c|}{}                                 & \textbf{Avg}            & {\underline{72.64}}    & 54.99       & \textbf{72.65} & 67.26                                                    & 64.21          & 64.87                                                      & 39.33                                                   & 40.37                                                   & \textbf{50.80}                                                 & \underline{49.84}                                                 \\ \midrule
\multicolumn{1}{c|}{\multirow{5}{*}{Game Rule}}       & IF                      & \textbf{39.75} & 24.25       & {\underline{39.19}}    & 30.10                                                    & 26.59          & 24.26                                                      & 14.75                                                   & \textbf{22.75}                                                   & 16.29                                                 & \underline{19.30}                                                 \\
\multicolumn{1}{c|}{}                                 & VC                      & 51.45          & 36.33       & {\underline{72.33}}    & 67.09                                                    & \textbf{72.71} & 68.79                                                      & 40.07                                                   & \underline{55.29}                                                  & \textbf{64.56}                                                 & 37.52                                                 \\
\multicolumn{1}{c|}{}                                 & VF                      & 77.95          & 59.15       & \textbf{88.18} & 80.59                                                    & {\underline{86.28}}    & 88.39                                                      & 59.13                                                   & 69.45                                                   & \textbf{80.13}                                                 & \underline{72.20}                                                 \\
\multicolumn{1}{c|}{}                                 & RC                      & {\underline{17.70}}    & 8.17        & \textbf{19.97} & 13.06                                                    & 15.45          & 15.61                                                      & 6.98                                                    & 7.56                                                    & \textbf{14.12}                                                 & \underline{10.48}                                                 \\
\multicolumn{1}{c|}{}                                 & \textbf{Avg}            & 46.71          & 31.98       & \textbf{54.92} & 47.71                                                    & {\underline{50.26}}    & 49.26                                                      & 30.23                                                   & 38.76                                                   & \textbf{43.77}                                                 & \underline{34.88}                                                 \\ \midrule
\multicolumn{1}{c|}{\multirow{5}{*}{Semantics Rule}}  & IF                      & \textbf{71.83} & 56.44       & {\underline{68.12}}    & 65.08                                                    & 59.91          & 61.28                                                      & 38.51                                                   & 46.06                                                   & \textbf{48.77}                                                 & \underline{46.27}                                                 \\
\multicolumn{1}{c|}{}                                 & VC                      & \textbf{92.65} & {\underline{91.18}} & 90.85          & {\underline{91.18}}                                              & 87.33          & 87.67                                                      & 80.39                                                   & 75.82                                                   & \textbf{82.35}                                                 & \underline{80.72}                                                 \\
\multicolumn{1}{c|}{}                                 & VF                      & \textbf{91.62} & 82.50       & 83.43          & {\underline{89.02}}                                              & 82.19          & 84.55                                                      & 79.17                                                   & 70.69                                                   & \underline{82.70}                                                 & \textbf{83.09}                                                 \\
\multicolumn{1}{c|}{}                                 & RC                      & \textbf{67.57} & 44.13       & 53.69          & {\underline{56.80}}                                              & 49.95          & 49.42                                                      & 32.01                                                   & 37.34                                                   & \underline{37.73}                                                 & \textbf{38.40}                                                 \\
\multicolumn{1}{c|}{}                                 & \textbf{Avg}            & \textbf{80.92} & 68.56       & 74.02          & {\underline{75.52}}                                              & 69.84          & 70.73                                                      & 57.52                                                   & 57.48                                                   & \textbf{62.89}                                                 & \underline{62.12}                                                 \\ \midrule
\multicolumn{1}{c|}{\multirow{5}{*}{Hypothesis Rule}} & IF                      & \textbf{86.97} & 58.55       & 72.44          & {\underline{80.13}}                                              & 71.93          & 64.32                                                      & 44.44                                                   & 41.45                                                   & \underline{61.11}                                                 & \textbf{61.75}                                                 \\
\multicolumn{1}{c|}{}                                 & VC                      & \textbf{85.90} & 64.32       & 77.35          & {\underline{81.62}}                                              & 66.45          & 67.74                                                      & 51.92                                                   & 50.43                                                   & \textbf{64.74}                                                 & \underline{55.56}                                                 \\
\multicolumn{1}{c|}{}                                 & VF                      & \textbf{92.20} & 81.54       & 82.50          & {\underline{85.73}}                                              & 76.86          & 79.66                                                      & 73.89                                                   & 63.8                                                    & \textbf{77.03}                                                 & \underline{75.17}                                                 \\
\multicolumn{1}{c|}{}                                 & RC                      & \textbf{46.79} & 12.50       & 41.35          & {\underline{46.69}}                                              & 18.31          & 28.31                                                      & 9.62                                                    & 11.00                                                   & \underline{12.93}                                                 & \textbf{17.84}                                                 \\
\multicolumn{1}{c|}{}                                 & \textbf{Avg}            & \textbf{77.96} & 54.23       & 68.41          & {\underline{73.54}}                                              & 58.39          & 60.01                                                      & 44.97                                                   & 41.67                                                   & \textbf{53.95}                                                 & \underline{52.58}                                                 \\ \midrule
\multicolumn{1}{c|}{\multirow{5}{*}{Humanity Rule}}   & IF                      & {\underline{79.90}}    & 53.46       & \textbf{80.04} & 72.87                                                    & 63.28          & 68.93                                                      & 46.56                                                   & 42.32                                                   & \underline{49.76}                                                 & \textbf{52.28}                                                 \\
\multicolumn{1}{c|}{}                                 & VC                      & {\underline{87.37}}    & 73.10       & \textbf{88.06} & 84.25                                                    & 79.83          & 83.13                                                      & 70.60                                                    & 54.23                                                   & \textbf{72.34}                                                 & \underline{70.47}                                                 \\
\multicolumn{1}{c|}{}                                 & VF                      & \textbf{94.49} & 84.38       & 88.08          & {\underline{89.65}}                                              & 83.90          & 88.52                                                      & 80.94                                                   & 67.76                                                   & \textbf{83.15}                                                 & \underline{82.32}                                                 \\
\multicolumn{1}{c|}{}                                 & RC                      & \textbf{61.23} & 35.23       & {\underline{56.78}}    & 50.63                                                    & 33.41          & 38.75                                                      & 27.78                                                   & 20.60                                                   & \textbf{30.21}                                                 & \underline{29.21}                                                 \\
\multicolumn{1}{c|}{}                                 & \textbf{Avg}            & \textbf{80.75} & 61.54       & {\underline{78.24}}    & 74.35                                                    & 65.10          & 69.83                                                      & 56.47                                                   & 46.23                                                   & \textbf{58.86}                                                 & \underline{58.57}                                                 \\ \midrule
\multicolumn{1}{c|}{\multirow{4}{*}{Vision Rule}}     & VC                      & 59.53          & 46.19       & 57.77          & 56.14                                                    & \textbf{70.04} & {\underline{61.86}}                                                & 43.49                                                   & 24.79                                                   & \textbf{59.03}                                                 & \underline{51.26}                                                 \\
\multicolumn{1}{c|}{}                                 & VF                      & {\underline{72.67}}    & 57.63       & 57.77          & 71.61                                                    & 68.32          & \textbf{76.06}                                             & \underline{52.94}                                                   & 29.41                                                   & \textbf{65.55}                                                 & 49.58                                                 \\
\multicolumn{1}{c|}{}                                 & RC                      & \textbf{48.94} & 30.58       & 28.50          & 40.47                                                    & {\underline{42.24}}    & 41.74                                                      & 18.91                                                   & 14.78                                                   & \textbf{29.34}                                                 & \underline{23.25}                                                 \\
\multicolumn{1}{c|}{}                                 & \textbf{Avg}            & \textbf{60.38} & 44.80       & 48.02          & 56.07                                                    & {\underline{60.20}}    & 59.89                                                      & 38.45                                                   & 22.99                                                   & \textbf{51.31}                                                 & \underline{41.36}                                                 \\ \midrule
\multicolumn{1}{c|}{\multirow{5}{*}{Average}}         & IF                      & \textbf{68.70}  & 46.97       & {\underline{65.16}}    & 61.06                                                    & 55.82          & 55.53                                                      & 33.84                                                   & 36.11                                                   & \underline{42.62}                                                 & \textbf{43.08}                                                 \\
\multicolumn{1}{c|}{}                                 & VC                      & 76.68          & 64.07       & \textbf{79.06} & {\underline{76.84}}                                              & 76.14          & 75.03                                                      & 55.82                                                   & 51.57                                                   & \textbf{68.59}                                                 & \underline{60.21}                                                 \\
\multicolumn{1}{c|}{}                                 & VF                      & \textbf{86.72} & 74.59       & 81.58          & {\underline{84.39}}                                              & 80.48          & 84.14                                                      & 69.56                                                   & 62.01                                                   & \textbf{78.15}                                                 & \underline{74.05}                                                 \\
\multicolumn{1}{c|}{}                                 & RC                      & \textbf{48.87} & 25.46       & 41.23          & {\underline{41.51}}                                              & 32.17          & 34.30                                                       & 17.99                                                   & 17.50                                                    & \textbf{23.58}                                                 & \underline{22.51}                                                 \\
\multicolumn{1}{c|}{}                                 & \textbf{Avg}            & \textbf{70.24} & 52.77       & {\underline{66.76}}    & 65.95                                                    & 61.15          & 62.25                                                      & 44.30                                                    & 41.80                                                    & \textbf{53.24}                                                 & \underline{49.96}                                                 \\ \midrule
\multicolumn{2}{c|}{Win Rate}                                                   & \textbf{0.397} & 0.186       & {\underline{0.340}}    & 0.300                                                    & 0.257          & 0.267                                                      & 0.151                                                   & 0.151                                                   & \textbf{0.193}                                                 & \underline{0.162}                                                 \\ \bottomrule
\end{tabular}
\vspace{-3pt}
\end{table*}

To ensure the quality of the checklists, we perform a careful manual verification for each sample.
%
%
We instruct human annotators to review each checklist question and make revisions based on accuracy, clarity, consistency, and completeness.
After manual verification, we finalize 6,500 checklist questions for 622 samples, ensuring comprehensive coverage across all four evaluation metrics.

%% file: sec/05_exp.tex
\section{Experiments and Evaluations}
\label{sec:exp}

\textbf{General Settings}.
%
We evaluate the rule-based reasoning capabilities of 10 state-of-the-art video models on \benchmark, including six closed-source models: Veo3.1~\cite{GoogleDeepMind2025Veo3}, Veo2~\cite{google-veo2-2025}, Sora2~\cite{openai2025sora2}, PixelVerse-V5~\cite{pixverse-v5-2025}, Wan2.5~\cite{wan2.5-model2025}, and Seedance1.0-pro~\cite{gao2025seedance}, and four open-source models: HunyuanVideo~\cite{kong2024hunyuanvideo}, CogVideoX1.5-5B~\cite{yang2024cogvideox}, Wan2.1-14B~\cite{wan2025}, and Wan2.2-A14B~\cite{wan2025}.
Notably, the four open-source models provide separate T2V and I2V versions, whose results are reported as a unified entry in our experiments. 
%

We use the default implementations of these video models, with detailed configurations provided in the Appendix.

\textbf{Metrics}. 
To enable straightforward comparison, we normalized all scores to a 100-point scale.
In particular, we map \textit{good}, \textit{medium}, and \textit{bad} ratings to 100, 50, and 0, respectively.
For each checklist, the score of each metric is computed as the average rating across all associated questions.
To obtain the overall score, we assign equal weights to the four metrics and compute their mean score.

\begin{table}
  \caption{Human alignment comparison across different MLLMs.
  }
  \label{tab:human_align}
  \centering
  \footnotesize
  \setlength\tabcolsep{8.8pt}
  \begin{tabular}{cc|cccc}
    \toprule
    \multicolumn{2}{c|}{\textbf{Model}} & \textbf{KRCC} & \textbf{SRCC} & \textbf{PLCC} & \textbf{ACC} \\ \midrule
    \multicolumn{2}{c|}{Claude-haiku-4-5} & 0.2941 & 0.5128 & 0.5199 & 0.6728 \\
    \multicolumn{2}{c|}{Claude-sonnet-4-5} & 0.3143 & 0.5356 & 0.5442 & 0.7036 \\
    \multicolumn{2}{c|}{Doubao-seed-1-6} & 0.1892 & 0.4827 & 0.5157 & 0.7632 \\
    \multicolumn{2}{c|}{Gemini-2.5-flash} & 0.3436 & 0.5976 & 0.6023 & 0.7368 \\
    \multicolumn{2}{c|}{Gemini-2.5-pro} & 0.3586 & 0.6073 & 0.6112 & 0.7208 \\
    \multicolumn{2}{c|}{Grok-4} & 0.3522 & 0.5504 & 0.5612 & 0.6820 \\
    \multicolumn{2}{c|}{GPT-4o} & 0.3044 & 0.5564 & 0.5615 & 0.7282 \\
    \multicolumn{2}{c|}{GPT-5} & {\underline{0.4204}} & {\underline{0.7295}} & {\underline{0.7289}} & {\underline{0.7995}} \\
    \multicolumn{2}{c|}{GPT-o3} & \textbf{0.4622} & \textbf{0.8011} & \textbf{0.8042} & \textbf{0.8512} \\
    \bottomrule
  \end{tabular}
\vspace{-15pt}
\end{table}

\begin{figure*}[t]
  \centerline{\includegraphics[width=\textwidth]{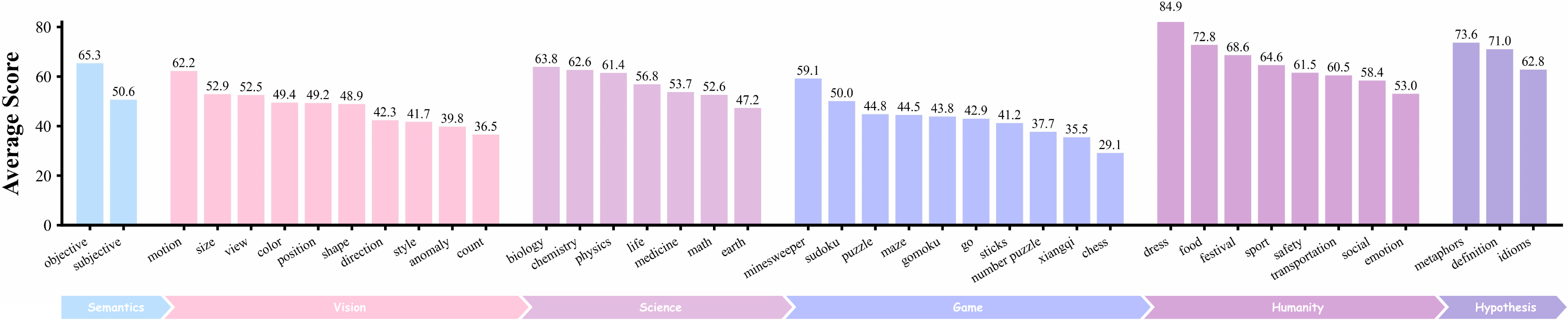}} 
  \caption{Average performance of video generation models across different tasks on \benchmark.
  Video models generally perform best on tasks in Humanity and Hypothesis, while showing lower performance on Vision and Game categories. 
  }
  
  \label{fig:task_result}
\end{figure*}

\subsection{Main Results}
We evaluate the performance of 10 state-of-the-art video generation models across six rule categories under four evaluation metrics. 
Our quantitative results in~\cref{tab:main_result} demonstrate that \benchmark effectively reveals fine-grained differences across reasoning capabilities and evaluation metrics.
We summarize the key observations below.

\textbf{Observation 1. Video models exhibit limited performance on the Rule Coherence metric compared to other evaluation metrics.} 
As shown in~\cref{tab:main_result}, all video models achieve their lowest performance on the Rule Coherence metric, highlighting the significant challenge of rule-based reasoning for current generation models.
Moreover, we observe a consistent decline in scores as the metrics reflect more complex cognitive skills.
For instance, Veo3.1 obtains an average score of approximately 80 on the perception-based metrics, including Visual Consistency and Visual Fidelity.
The score drops to 68.7 on the Instruction Following metric, which evaluates visual understanding, and further decreases to 48.87 on the Rule Coherence metric.
In addition, open-source models generally exhibit lower reasoning capabilities compared to closed-source models, suggesting that higher-quality training data and larger model capacity can enhance rule-based reasoning ability.

\textbf{Observation 2. Video generation models exhibit diverse performance patterns across different categories.} 
As illustrated in~\cref{tab:main_result}, video models achieve the highest scores on the Humanity and Semantics categories, with Veo3.1 reaching an average of 80.
This indicates that current models exhibit strong generative competence in the social domain, demonstrating a solid understanding of human customs, language, and behaviors.
In contrast, all models achieve an average Rule Coherence score below 15 on the Game category, suggesting limited generalization ability in customized, strategy-driven scenarios.
%
%
As shown in~\cref{fig:task_result}, we further analyze the average performance of different video models across tasks.
We observe that models also exhibit considerable intra-category variations in reasoning ability.
For example, under the Humanity category, video models generally achieve higher scores on the dress task than on other tasks, reaching an average of 84.9, which can be attributed to the prevalence of clothing-related cues in human-centric scenes, thus enabling video models to generalize dress-related rules through training.

\begin{table}
  \caption{Effect of Prompt Enhancement (PE) on Rule Coherence.
    (Game category is excluded since most samples rely heavily on ground truth images, thus less affected by PE).
    }
  \label{tab:enhance_prompt}
  \centering
  \footnotesize
  \setlength\tabcolsep{3.9pt}
    \begin{tabular}{c|ccccc}
    \toprule
    \textbf{Method}      & \textbf{Science} & \textbf{Vision} & \textbf{Semantics} & \textbf{Hypothesis} & \textbf{Humanity} \\ \midrule
    Veo3.1      & 50.97   & 48.94  & 67.57     & 46.79      & 61.23    \\
    Veo3.1+PE   & 62.62   & 52.61  & 72.19     & 58.12      & 67.40    \\
    \noalign{\kern-\tabcolsep}
    \rowcolor{gray!20}
    \noalign{\kern\tabcolsep}
    $\bigtriangleup$ & +9.65   & +3.67  & +4.62     & +11.33     & +6.17    \\ \midrule
    Sora2       & 47.09   & 28.50  & 53.69     & 41.35      & 56.78    \\
    Sora2+PE    & 55.45   & 36.06  & 63.37     & 63.35      & 65.15    \\
    \rowcolor{gray!20}
    $\bigtriangleup$ & +8.36   & +7.56  & +9.68     & +22.0      & +8.37    \\ \bottomrule
    \end{tabular}
\vspace{-15pt}
\end{table}

\textbf{Observation 3. Video models struggle with visual understanding and reasoning.}
As shown in~\cref{tab:main_result} and~\cref{fig:task_result}, generative models perform worst on the Game and Vision categories, both of which consist entirely of I2V instances.
This indicates that, beyond their limited capabilities of rule-based reasoning, the models also exhibit insufficient reasoning of image content, which further constrains their capability to generate high-quality videos.
To obtain deeper insight into this limitation, we analyze the result across task types and find that video generation models achieve an average score of 65.55 on T2V tasks, significantly higher than 48.56 on I2V tasks, indicating that an increase in input modalities leads to a sharp decline in reasoning performance.

\begin{figure*}[t]
  \centerline{\includegraphics[width=\textwidth]{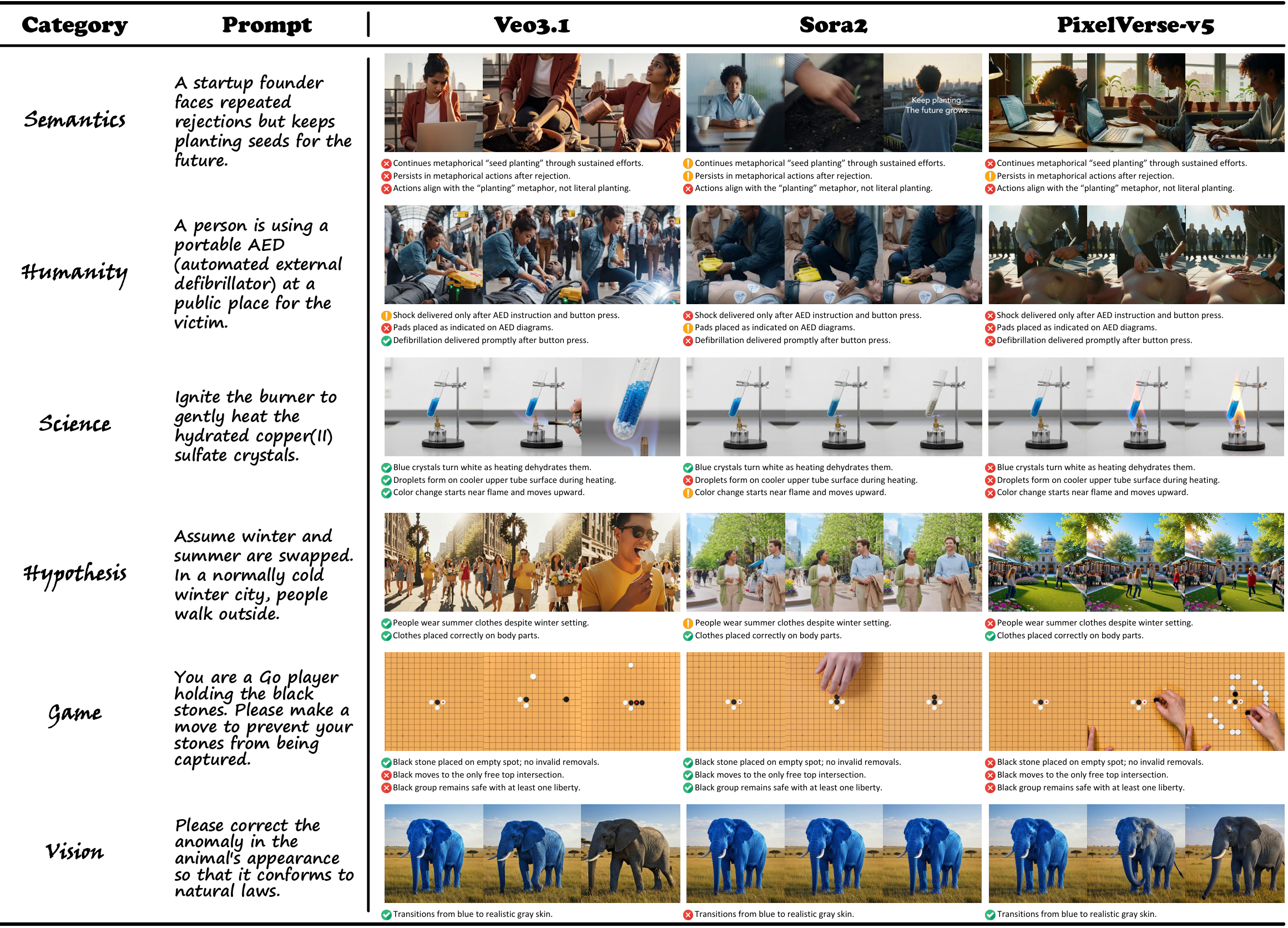}} 
  \caption{Case studies on three closed-source models across six rule categories.
            Each sample is provided with the Rule Coherence aspects derived from the checklist questions.
            The three video models exhibit varying performance across different instances.
            }
  \label{fig:case_study}
\end{figure*}

\subsection{Analysis}
\textbf{Impact on Prompt Enhancement}. To analyze the impact of prompt enhancement at test times, we use GPT-o3 to generate an enhanced prompt based on the \textit{prompt} and \textit{implicit explanation} of each sample, explicitly incorporating the expected outcomes.
As shown in~\cref{tab:enhance_prompt}, prompt enhancement substantially improves the performance of Veo3.1 and Sora2 on Rule Coherence.
However, these results also suggest that even with explicit guidance on the expected phenomena, current video models still struggle to generate rule-consistent video clips based on their reasoning abilities, revealing a remaining gap toward zero-shot reasoners.

\textbf{Case study}.
~\cref{fig:case_study} presents qualitative case studies of video models' performance on \benchmark across different rule categories.
In row \#1, all models misinterpret the metaphor ``\textit{planting seeds for the future}'' literally, depicting the sowing and watering scenes.
In row \#2, Veo3.1 fails to infer the correct placement of AED electrodes, while the other two models show limited understanding of the device’s function.
In row \#3, Veo3.1 and Sora2 demonstrate scientific reasoning abilities, with Veo3.1 exhibiting finer-grained details.
Meanwhile, in row \#4, all models correctly infer the summer setting, yet their abilities to reason about human clothing vary.
In row \#5, Sora2 successfully performs game-strategy reasoning, whereas the other models fail to comprehend the task instruction.
Finally, in row \#6, Veo3.1 and PixelVerse-v5 effectively detect and rectify visual anomalies, while Sora2 struggles to reason over the visual context.
These results show notable differences in reasoning competence across models, reflecting their varying abilities to integrate conceptual cues.

\textbf{Human alignment validation}.
To evaluate the reliability of our evaluation protocols, we conduct a human preference alignment experiment.
First, we randomly select 80 generated videos along with the corresponding 813 checklist questions.
Second, human annotators are invited to answer each question based on the video content.
Third, we evaluate the consistency between outputs of different closed-source MLLMs and human annotations, reporting Kendall Rank Correlation Coefficient (KRCC), Spearman Rank Correlation Coefficient (SRCC), Pearson Linear Correlation Coefficient (PLCC), and the overall Accuracy (ACC).
As shown in~\cref{tab:human_align}, all models demonstrate strong video understanding capabilities.
Among them, GPT-o3 achieves the highest accuracy of 0.8512 and outperforms other MLLMs across all three correlation metrics, demonstrating the reliability of using GPT-o3 as the evaluator.

%% file: sec/06_conclusion.tex
\section{Conclusion}
\label{sec:conclusion}

In this paper, we introduce \benchmark, a comprehensive benchmark designed to evaluate the rule-based reasoning abilities of video generation models.
\benchmark comprises 622 high-quality instances spanning 40 tasks across six rule categories, addressing the critical need for fine-grained evaluation of reasoning capabilities in video models.
To ensure objective evaluation, we design a checklist-based protocol from four evaluation metrics.
Finally, we conduct extensive experiments on 10 state-of-the-art video generation models and provide detailed analyses based on the results.
We hope our work will provide valuable insights for advancing reasoning-aware video generation towards vision foundation intelligence.

%% file: sec/07_supp.tex
\renewcommand\thefigure{A\arabic{figure}}
\renewcommand\thetable{A\arabic{table}}  
\renewcommand\theequation{A\arabic{equation}}
\setcounter{equation}{0}
\setcounter{table}{0}
\setcounter{figure}{0}

\clearpage
\maketitlesupplementary
\appendix

\section*{Appendix Contents}
The Appendix of \benchmark is structured as follows:
\begin{itemize}
\setlength{\leftskip}{1em}
    \item \cref{supp_sec:detail}: Detailed task explanations. 
    \item \cref{supp_sec:imp_detail}: Implementation settings and configurations of different video generation models.
    \item \cref{supp_sec:explore}: Explorations on Video-to-Video tasks.
    \item \cref{supp_sec:user_study}: More detailed configuration and experimental results of human annotation and user study.
    \item \cref{supp_sec:prompt}: Prompts used for evaluation and generation.
    \item \cref{supp_sec:exp_res}: Experimental results across tasks.
    \item \cref{supp_sec:vis}: More visualization results.
    \item \cref{supp_sec:limitation}: Limitations and future works.
    \item \cref{supp_sec:social_impact}: Potential social impact of \benchmark.
\end{itemize}

\input{supp_secs/01_detail}

\input{supp_secs/02_implementation_detail}
\input{supp_secs/03_v2v_res}
\input{supp_secs/04_user_study}
\input{supp_secs/05_gen_eval_prompt}
\input{supp_secs/06_exp_res}
\input{supp_secs/07_vis}
\input{supp_secs/08_limitation}
\input{supp_secs/09_social_impact}

\input{supp_secs/open_table}
\input{supp_secs/open_figure}
\input{supp_secs/open_prompt}

%% file: supp_secs/01_detail.tex
\section{Detailed Tasks Explanations}
\label{supp_sec:detail}
\benchmark evaluates the reasoning abilities of the state-of-the-art video generation models across six rule categories: Vision, Science, Semantics, Hypothesis, Game, and Humanity.
To enable a more fine-grained assessment, we design a suite of 40 tasks distributed over these categories, spanning a broad spectrum of scenarios across the \textit{nature}, \textit{society}, and \textit{virtuality} domains.
In the following section, we comprehensively define each task.

\subsection{Vision Rule}
The Vision category comprises 10 distinct tasks, covering physical, spatial, and temporal visual attributes.
Tasks in this category are designed to evaluate whether video generation models can accurately infer the visual composition of the real world from a given image and generate videos that adhere to fundamental visual principles.

\textbf{Anomaly} task focuses on visual plausibility reasoning.
Given an image containing a clear visual anomaly, generative models are required to infer the correct appearance of the object based on implicit visual rules and generate videos that restore visual coherence.
For example, when the input image depicts a cat with four tails, the model should reason that a cat normally has only one tail and accordingly correct the anomaly in the generated video clip.

\textbf{Color} task focuses on color reasoning and manipulation.
Given an input image, generative models are required to reason about the color attributes of specific objects and modify them while preserving spatial-temporal coherence.

\textbf{Size} task targets object-scale adjustment and proportional reasoning.
Given an instruction to resize a particular object, the model should modify its scale appropriately while preserving scene geometry and physical plausibility.

\textbf{Count} task evaluates numerical reasoning and controllability.
Models are required to adjust the number of instances of a given object according to the instruction while preserving spatial realism and temporal coherence.

\textbf{Direction} task focuses on directional reasoning and motion control.
Given an input image and a directional instruction, generative models are required to adjusting the object's facing and orientation, ensuring temporal coherence.

\textbf{Shape} task evaluates geometric reasoning and controllability.
Video models are required to reason about the specified object's geometry and modify its structure accordingly.

\textbf{Position} task evaluates the reasoning ability of video models on spatial attributes.
The model needs to reason about the object's location and relocate it accordingly.

\textbf{View} task focuses on viewpoint reasoning and perspective adaptation.
Given a target view specification, video models should generate videos from the new viewpoint. 

\textbf{Motion} task targets dynamic prediction and temporal reasoning.
Models are required to infer plausible motion trajectories from a given static or partial motion input and predict temporally coherent frames.
For instance, when observing a basketball player suspended in mid-air, the model should predict and generate the subsequent motion that reflects a realistic landing and continuation of the action.

\textbf{Style} task focuses on appearance reasoning and stylistic transformation.
Given an input image and a style transformation instruction, video models are required to apply the target style while preserving object identity.

\subsection{Science Rule}
The Science category consists of 10 tasks spanning multiple disciplines within the natural sciences.
These tasks are designed to evaluate whether video generation models have acquired a fundamental understanding of natural laws and scientific principles, enabling them to accurately simulate experimental phenomena, biological behaviors, and other science-driven processes in the real world.

\textbf{Physics} task focuses on scenes that involve fundamental physical principles, covering multiple domains such as mechanics, optics, and electromagnetism.
Video models are required to predict physically consistent phenomena based on the implicit laws embedded within each sample.

\textbf{Chemistry} task focuses on fundamental chemical reaction principles and their observational experimental manifestations.
Video generation models are required to infer plausible reaction processes and products by reasoning over the chemical rules, given the reactants and conditions.

\textbf{Biology} task focuses on fundamental biological principles.
Video models are required to infer plausible experimental outcomes based on organism-specific characteristics, covering a variety of scenarios such as controlled biological experiments, genetics, and conditioned reflexes.

\textbf{Earth} task focuses on geographical and environmental phenomena driven by factors such as location, climate variation, and seasonal cycles.
Video generation models are required to infer long-term scene evolution according to underlying geographical principles and reflect these changes coherently within the generated video clip.

\textbf{Math} task focuses on the application of fundamental mathematical principles.
Given a mathematical problem, video generation models are required to apply the relevant principles, derive the solution step by step, and visually illustrate the reasoning process clearly and coherently.

\textbf{Medicine} task focuses on the correct application and demonstration of medical protocols.
Given a clinical scenario, video generation models are required to follow the specified procedural standards, such as venipuncture guidelines, the seven-step handwashing protocol, or basic surgical procedures, and accurately depict these procedures.

\textbf{Life} task focuses on fundamental principles of animal behavior.
Video generation models are required to infer plausible behavioral responses when experimental disturbances occur, reasoning according to species-specific behavioral patterns and ecological logic.

\subsection{Hypothesis Rule}
The Hypothesis category includes two tasks that encompass diverse scenarios, designed to evaluate whether video generation models can reason according to newly defined rules and assumptions beyond real-world rules.
Unlike other categories, the governing rules here are explicitly provided in the prompt, requiring the model to perform zero-shot reasoning under newly introduced conditions.

\textbf{Subjective Scenario} task focuses on human-defined hypothetical worlds.
These scenarios violate real-world logic, requiring video generation models to reason about state changes according to the provided subjective rules.
For example, given the rule ``\textit{lying causes the nose to grow}'', the model must recognize a lying event in the scene and apply the specified rule to generate a video in which the character's nose lengthens accordingly. 

\textbf{Objective Law} task focuses on modifying objective real-world principles.
Unlike subjective scenarios, Objective Law tasks remain grounded in real physical principles but introduce deliberate modifications to them.
Video models are required to abandon the original rule, adopt the newly specified one, and perform reasoning under the modified physical dynamics.
For example, given the instruction “assume the density order is reversed,” the model must reason according to the new law and generate outcomes that contradict the behavior dictated by real-world physics.

\subsection{Game Rule}
The Game category comprises 10 tasks that cover a wide range of logical reasoning and game scenarios.
These tasks are designed to evaluate whether video generation models can reason according to game strategies, execute rational moves, and ultimately achieve victory.
To ensure controllable complexity, each sample is constrained to the minimal number of required moves and simplified difficulty settings.

\textbf{Chinese Chess} task focuses on one-move checkmate scenarios.
Models are required to generate a single, legally valid move that adheres to the piece-specific movement rules and delivers a decisive checkmate against the opponent’s General, ensuring that the resulting board state is both legally valid and tactically forced.

\textbf{Gomoku} task focuses on fundamental offensive and defensive patterns such as open-three formations, double-three configurations, or immediate winning opportunities.
Models must infer the optimal placement to either complete a five-in-a-row or block the opponent's winning threat.

\textbf{Go} task focuses on one-move capture scenarios.
Models are required to reason over the current board configuration and place a single stone that removes one or more opponent groups that are in atari with only one remaining liberty, thereby executing a valid and tactically justified capture.

\textbf{Sudoku} task evaluates logical deduction under strict numerical constraints.
Given a partially filled grid, models must infer the correct number placement that simultaneously satisfies row, column, and subgrid consistency rules.

\textbf{Chess} task targets tactical positions where a checkmate can be achieved in the next move.
Models must analyze the board configuration, identify forced-mate patterns, and generate the correct move that results in a checkmate.

\textbf{Minesweeper} task focuses on number-based spatial constraints.
Models must analyze the revealed mine counts and infer which unrevealed tiles must contain mines with logical certainty, subsequently producing a move that marks all deterministically identifiable mine locations.

\textbf{Maze} task evaluates pathfinding and spatial reasoning under structural constraints.
Models are required to generate a sequence of movements that leads the agent from the starting point to the goal while avoiding dead ends.

\textbf{Puzzle} task focuses on identifying spatial correspondences and assembling fragmented pieces into a unified structure.
Models must infer the correct arrangement of puzzle pieces by reasoning about shape compatibility and boundary alignment to form a coherent final pattern.

\textbf{Number Sliding} task evaluates sequential planning under movement constraints.
Models must generate a series of sliding operations that reposition numbered tiles toward the target configuration while preserving the adjacency rules.

\textbf{Matchsticks} task focuses on geometric and arithmetic reasoning under limited move constraints.
Models must determine which matchsticks to reposition in order to form a valid equation or geometric configuration.

\subsection{Semantics Rule}
The Semantics category comprises 3 tasks, designed to evaluate whether video generation models can infer high-level semantic information from contextual cues and generate videos that accurately reflect the intended visual meaning.

\textbf{Definition} task focuses on the accurate illustration of a provided term or concept. 
Models are required to infer the essential semantic attributes and contextual nuances of the term and produce visual content that aligns with its correct definition rather than surface-level associations.

\textbf{Metaphor} task focuses on the interpretation of metaphorical expressions
Given a metaphorical phrase, the models are required to generate a video that represents the underlying conceptual meaning, rather than a literal depiction of the words, demonstrating understanding of figurative language and abstract semantic mapping.

\textbf{Idiom} task focuses on culturally conventional idiomatic expressions.
Models need to avoid literal interpretations and instead generate scenes that correctly express the idiom’s figurative semantic content, reflecting understanding of linguistic conventions and contextualized meaning.

\subsection{Humanity Rule}
The Humanity category includes 8 tasks that encompass diverse social scenarios and human conventions.
These tasks are designed to evaluate whether video generation models can reason about societal norms, cultural practices, and customs to generate plausible human behaviors.

\textbf{Dress} task focuses on culturally appropriate clothing choices and outfit consistency.
Models are required to generate videos where characters wear attire suitable for the given context, season, or activity.

\textbf{Food} task examines dietary norms and context-aware food selection.
Models must generate scenes where the food type, preparation style, or eating behavior aligns with common culinary conventions and the specified scenario.

\textbf{Emotion} task evaluates the understanding of human affect and expressive behavior.
Models are required to generate facial expressions, gestures, and body movements that reflect the specified emotional state in a natural manner.

\textbf{Festival} task focuses on cultural celebrations and associated symbolic practices.
Models must incorporate appropriate festival elements, such as decorations, attire, or rituals, corresponding to the designated cultural event.

\textbf{Safety} task assesses whether models can adhere to basic safety norms in daily life.
Given specific situations, models must avoid unsafe behaviors and instead generate actions aligned with conventional safety practices.

\textbf{Social} task focuses on everyday social conventions.
Models are required to generate actions that adhere to common social norms, engaging in other routine prosocial behaviors that reflect cultural expectations.

\textbf{Sports} task targets sport-specific conventions.
Models must generate videos where human movements, equipment usage, and gameplay dynamics align with the rules and common practices of the designated sport.

\textbf{Transportation} task focuses on concrete road-traffic rules.
Models are required to generate plausible interactions with real-world traffic systems, such as obeying traffic lights and road signs, or navigating multi-lane roads according to standard driving conventions.

%% file: supp_secs/02_implementation_detail.tex
\section{Implementation Detail}
\label{supp_sec:imp_detail}

\begin{table}
  \caption{Details of the configurations of generated videos.}
  \label{supp_tab:imp_detail}
  \centering
  \footnotesize
  \setlength\tabcolsep{5.7pt}
\begin{tabular}{ccccc}
\toprule
\textbf{Model}           & \textbf{Resolution} & \textbf{Frames} & \textbf{FPS} & \textbf{Durations (s)} \\ \midrule
\rowcolor{gray!20}
\multicolumn{5}{c}{\textbf{\textit{Closed-Source Model}}}                     \\ \midrule
Veo3.1          & 1280$\times$720   & 192    & 24  & 8.00          \\
Veo2            & 1280$\times$720   & 192    & 24  & 8.00          \\
Sora2           & 1280$\times$704   & 300    & 30  & 10.00         \\
PixelVerse-V5   & 1280$\times$720   & 121    & 24  & 5.04          \\
Seedance1.0-pro & 1248$\times$704   & 121    & 24  & 5.04          \\
Wan2.5          & 1280$\times$720   & 121    & 24  & 5.04          \\ \midrule
\rowcolor{gray!20}
\multicolumn{5}{c}{\textbf{\textit{Open-Source Model}}}                       \\ \midrule
CogVideoX1.5 5B & 1360$\times$768   & 161    & 16  & 10.06         \\
HuyuanVideo     & 832$\times$480    & 81     & 15  & 5.40          \\
Wan2.1 14B      & 832$\times$480    & 81     & 16  & 5.06          \\
Wan2.2 A14B     & 832$\times$480    & 81     & 15  & 5.40          \\ \bottomrule
\end{tabular}
\vspace{-15pt}
\end{table}

We follow the official configurations of different video generation models.
As summarized in~\cref{supp_tab:imp_detail}, the videos have a duration of 5$\sim$10 seconds and a resolution of either 720P or 480P.
Specifically, for open-source models, we implement them using the Diffusers framework~\cite{von-platen-etal-2022-diffusers}.

%% file: supp_secs/03_v2v_res.tex
\section{Explorations on Video-to-Video tasks}
\label{supp_sec:explore}
\begin{table*}[ht]
\centering
\setlength\tabcolsep{4.3pt}
\footnotesize
\caption{
    Human alignment comparison across MLLMs.
    KRCC, SRCC, PLCC, and ACC are reported across the four evaluation metrics.
}
\label{supp_tab:human_align}
\begin{tabular}{@{}c|cccc|cccc|cccc|cccc@{}}
\toprule
\multirow{2}{*}{\textbf{Model}} & \multicolumn{4}{c|}{\textbf{KRCC}}          & \multicolumn{4}{c|}{\textbf{SRCC}}          & \multicolumn{4}{c|}{\textbf{PLCC}}          & \multicolumn{4}{c}{\textbf{ACC}}           \\ \cmidrule(l){2-17} 
                       & IF     & RC     & VC     & VF      & IF     & RC     & VC     & VF      & IF     & RC     & VC     & VF      & IF     & RC     & VC     & VF     \\ \midrule
Claude-haiku-4-5       & .2655 & .3913 & .2912 & .1410  & .4186 & .5805 & .5139 & .3647  & .4384 & .5789 & .5028 & .3514  & .6456 & .6188 & .6312 & .7474 \\
Claude-sonnet-4-5      & .2998 & .3913 & .3375 & .1414  & .4852 & .5832 & .5809 & .3397  & .4909 & .5794 & .5640 & .3630  & .6519 & .6733 & .7188 & .7440 \\
Grok-4                 & .3656 & .4129 & .3874 & .2184  & .5680 & .6200 & .5990 & .4097  & .5798 & .6150 & .6042 & .4251  & .6603 & .7050 & .7006 & .6678 \\
doubao-seed-1-6        & .2540 & .3785 & .1387 & -.0031 & .5473 & .5521 & .3904 & -.0385 & .5569 & .5521 & .4143 & -.0385 & .6471 & .6875 & .6875 & .9259 \\
Gemini-2.5-flash       & .3725 & .4032 & .3238 & .1874  & .6021 & .6091 & .5647 & .4791  & .6113 & .6063 & .5649 & .4805  & .7025 & .7129 & .7250 & .7782 \\
Gemini-2.5-pro         & .4012 & .4426 & .3224 & .1708  & .6639 & .6926 & .5478 & .3843  & .6559 & .6921 & .5431 & .3974  & .7152 & .7525 & .6937 & .7167 \\
GPT-4o                 & .3336 & .3670 & .3064 & .1535  & .5869 & .5512 & .5661 & .4209  & .5777 & .5534 & .5516 & .4037  & .6962 & .6584 & .7125 & .8020 \\
GPT-5                  & .4598 & .4923 & .4152 & .1831  & .7495 & .7751 & .6898 & .5301  & .7309 & .7750 & .6869 & .5284  & .8038 & .7871 & .7625 & .8259 \\
GPT-o3                 & .5281 & .5218 & .4564 & .2419  & .8770 & .8188 & .8010 & .6315  & .8765 & .8171 & .8015 & .5996  & .8924 & .8218 & .8500 & .8498 \\ \bottomrule
\end{tabular}
\end{table*}

To provide a more systematic and comprehensive evaluation, we further explore the rule-based reasoning abilities of video generation models on the Video-to-Video (V2V) tasks.
Following the same data collection pipeline as in the I2V task, we manually curate web videos and use GPT-5 for captioning.
This web video set covers eight tasks from the Vision category and consists of 81 high-quality samples.
We leverage Wan2.1 14B to perform rule-based reasoning based on these V2V instances.
We find that in most cases, Wan2.1 14B struggles to perform plausible reasoning based on the prompt and input video, exhibiting relatively poor generative performance on V2V tasks, as illustrated in~\cref{supp_tab:v2v}
Surprisingly, the model achieves substantially higher scores on the Visual Consistency and Visual Fidelity metrics compared to I2V instances in the same task. 
We attribute this to the presence of the input video, which likely serves as a strong visual reference and encourages the model to align both visual quality and temporal consistency closely with the provided input.
However, in a few instances, as shown in~\cref{supp_fig:v2v}, Wan2.1 14B is able to produce results that meet expectations, although this capability is observed only on sporadic samples.

%% file: supp_secs/04_user_study.tex
\begin{table}
  \caption{Quantitative results of Wan2.1 14B for V2V tasks.}
  \label{supp_tab:v2v}
  \centering
  \footnotesize
  \setlength\tabcolsep{2.8pt}
  \begin{tabular}{@{}c|cccccccc@{}}
    \toprule
    Metrics & Color & Count & Direction & Position & Shape & Size  & Style & View  \\ \midrule
    RC      & 3.57  & 18.18 & 10.00     & 12.50    & 18.18 & 14.58 & 7.5   & 5.00  \\
    VC      & 42.86 & 22.73 & 55.00     & 71.67    & 31.82 & 62.5  & 52.5  & 42.50 \\
    VF      & 89.29 & 64.39 & 82.50     & 77.50    & 59.09 & 72.92 & 50.83 & 51.67 \\
    Avg     & 45.24 & 35.10 & 49.17     & 53.89    & 36.36 & 50.00 & 36.94 & 33.06 \\ \bottomrule
    \end{tabular}
\end{table}

\begin{figure}[t]
  \centering
  \includegraphics[width=\linewidth]{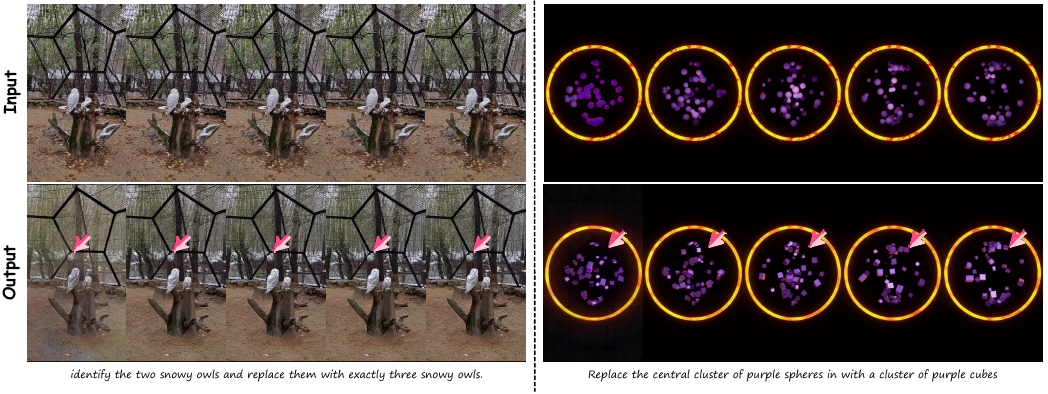}
  \caption{Qualitative results of Wan2.1 14B for V2V tasks.
  }
  \label{supp_fig:v2v}
  \vspace{-15pt}
\end{figure}

\section{Details of Human Annotation}
\label{supp_sec:user_study}

We conduct large-scale human annotation for two aspects: checklist validation and human alignment experiments.
In the checklist validation stage, human annotators are employed to assess the soundness of the proposed checklist and refine it when necessary.
In the human alignment experiment, annotations are collected to support a user study designed to evaluate the alignment between the output of GPT-o3 and human preferences.
The overall annotation process is collaboratively conducted by the authors.
In this section, we present the detailed annotation procedures and results.

\subsection{Checklist Validation}
To validate the effectiveness of the constructed checklist, we perform quality control for each individual checklist question.
For each sample, we provide detailed explanations of the associated metadata, such as the prompt and implicit explanation, and standardize the annotation requirements to ensure reliable judgments.
Each annotators are required to make revisions based on the following criteria:
\begin{itemize}
    \item \textit{Accuracy} evaluates whether the question accurately reflects the expected video content based on specific rules.
    \item \textit{Clarity} considers whether the checklist question is stated clearly and without ambiguity.
    \item \textit{Relevance} measures whether the question is relevant to the scenario and prompt of the given sample.
    \item \textit{Consistency} assesses whether the question type aligns with the corresponding evaluation dimension.
    \item \textit{Completeness} ensures whether the checklist covers all necessary aspects for thorough evaluation.
\end{itemize}

\subsection{User Study}

\begin{figure}[t]
  \centering
  \includegraphics[width=\linewidth]{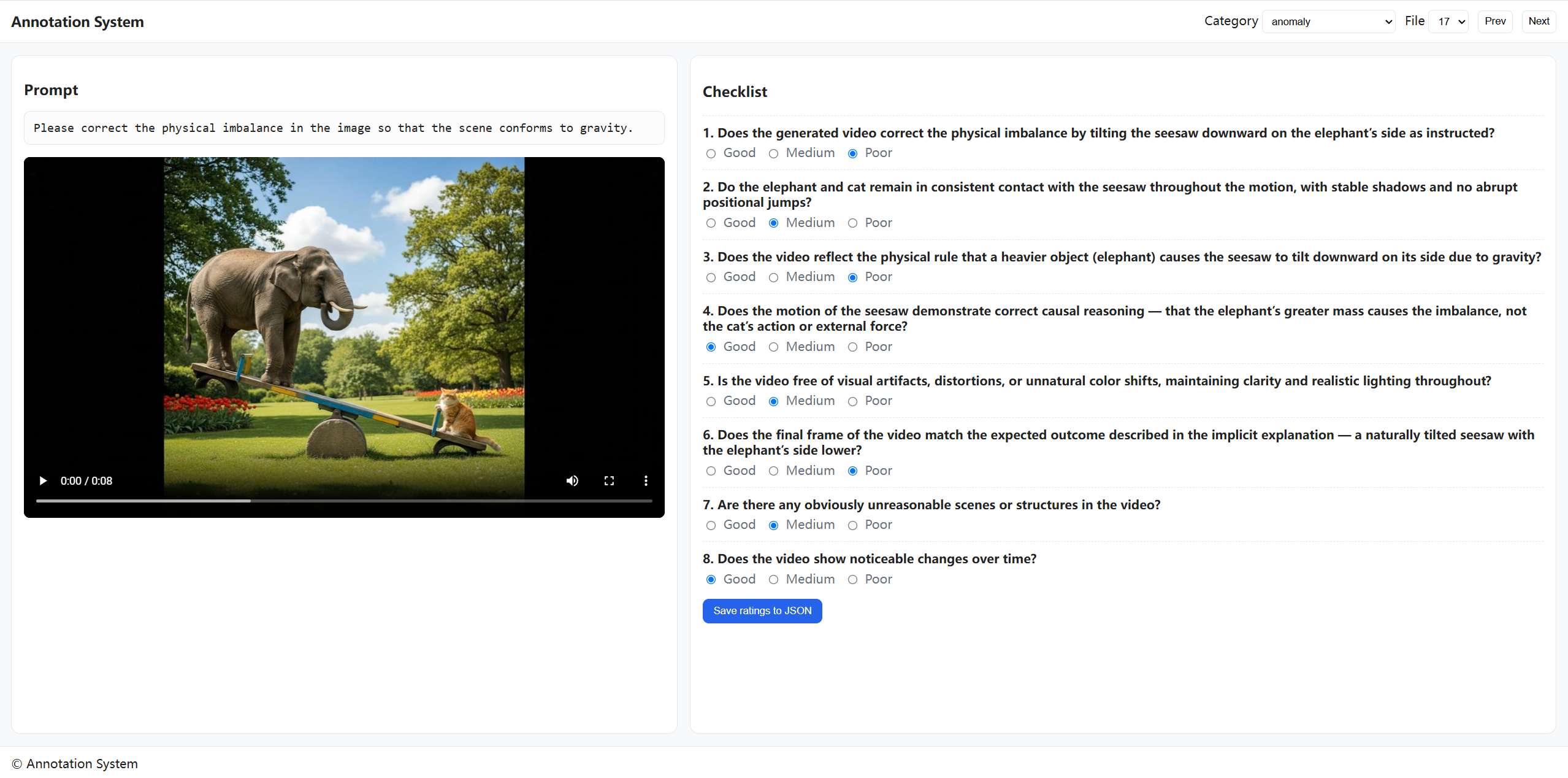}
  \caption{Annotation interface for human alignment evaluation.
  }
  \label{supp_fig:human}
  \vspace{-15pt}
\end{figure}

To verify the effectiveness of our GPT-o3-based automated evaluation, we conduct a human preference alignment experiment.
First, we randomly sample 80 generated videos along with their corresponding checklist questions.
Second, we develop an annotation interface to improve labeling efficiency.
As shown in~\cref{supp_fig:human}, annotators are provided with the generated video, the corresponding prompt, and the checklist questions.
Third, annotators need to complete each checklist question based on the visual content of the video.
After annotation, we aggregate all responses and compare human annotation with the output from state-of-the-art MLLMs, including Claude Haiku 4.5~\cite{claudehaiku45}, Claude Sonnet 4.5~\cite{claudesonnet45}, Doubao Seed 1.6~\cite{seed16}, Gemini 2.5 flash~\cite{comanici2025gemini}, Gemini 2.5 pro~\cite{comanici2025gemini}, Grok-4~\cite{grok4}, GPT-4o~\cite{hurst2024gpt}, GPT-5~\cite{openai2025gpt5}, and GPT-o3~\cite{openai2025o3}.
KRCC, SRCC, PLCC, and ACC are reported.

As shown in~\cref{supp_tab:human_align}, GPT-o3 demonstrates a significant performance advantage across all four evaluation metrics.
Its alignment with human preferences exceeds 85\%, indicating its effectiveness as an evaluator.

%% file: supp_secs/05_gen_eval_prompt.tex
\section{Prompt for Generation and Evaluation}
\label{supp_sec:prompt}
\cref{supp_fig:eval_prompt} illustrates the prompt used to evaluate the performance across four evaluation metrics of MLLMs.
\cref{supp_fig:checklist} presents the prompt for generating a checklist based on the provided metadata of the specific instance.
\cref{supp_fig:I2S} demonstrates the system prompt used for our Image-to-scenario pipeline. GPT-5 is required to generate the corresponding \textit{prompt} and \textit{implicit explanation} based on the instruction and provided image.
Finally, we use the prompt presented in~\cref{supp_fig:S2I} to generate \textit{initial scenario}, \textit{prompt}, and \textit{implicit explanation} according to the task property.

%% file: supp_secs/06_exp_res.tex
\section{Experimental Results per Task}
\label{supp_sec:exp_res}

We summarize the performance of different video generation models across the evaluation tasks.
As shown in~\cref{supp_tab:main_result}, Veo3.1 demonstrates impressive generative capability on multiple tasks, and closed-source models consistently outperform open-source models.
Moreover, we find that although prompt enhancement leads to observable performance gains, current video models still fall short of fully achieving reliable rule-based reasoning.

%% file: supp_secs/07_vis.tex
\section{More Visualization Results}
\label{supp_sec:vis}

Additional qualitative results are presented as follows:
\begin{itemize}
\setlength{\leftskip}{1em}
    \item \cref{supp_fig:anomaly}: Qualitative results of the Anomaly task. 
    \item \cref{supp_fig:safety}: Qualitative results of the Safety task.
    \item \cref{supp_fig:earth}: Qualitative results of the Earth task.
    \item \cref{supp_fig:life}: Qualitative results of the Life task.
    \item \cref{supp_fig:physics}: Qualitative results of the Physics task.
    \item \cref{supp_fig:chemistry}: Qualitative results of the Chemistry task.
    \item \cref{supp_fig:emotion}: Qualitative results of the Emotion task.
    \item \cref{supp_fig:color}: Qualitative results of the Color task.
\end{itemize}

%% file: supp_secs/08_limitation.tex
\section{Limitations and Future Works}
\label{supp_sec:limitation}
While \benchmark provides a comprehensive evaluation for the rule-based reasoning ability of video generation models, there are still many challenges:
\begin{itemize}
    \item Lack of a unified evaluation that jointly assesses T2V, I2V, and V2V tasks.
    Since most existing video models support either T2V and I2V jointly or V2V exclusively, we focus our evaluation on the T2V and V2V settings for consistency.
    Future work may extend this framework to all three task paradigms, enabling a unified assessment for the next-level video models that natively support T2V, I2V, and V2V in a comprehensive manner.
    \item Potential for both depth and breadth expansion. 
    While \benchmark covers a broad spectrum of rule-based reasoning scenarios, its breadth can be further extended to tasks such as GUI interaction, classical algorithms, and table reasoning, which are left for future work.
\end{itemize}

%% file: supp_secs/09_social_impact.tex
\section{Social Impacts}
\label{supp_sec:social_impact}
This work may positively influence the development of more reliable video generation by encouraging consistent rule-based reasoning, thereby supporting better downstream applications.
However, stronger generative abilities may also increase the risk of misuse that obeys ethical norms and legal requirements.
These risks highlight the importance of ethical deployment practices and appropriate oversight to ensure that progress in video generation technology benefits society while maintaining responsible use.

%% file: supp_secs/open_table.tex
\begin{table*}[ht]
\centering
\setlength\tabcolsep{3pt}
\footnotesize
\caption{
    Evaluation result across tasks in six rule categories. 
}
\label{supp_tab:main_result}
\begin{tabular}{@{}c|c|cccccc|cccc|cc@{}}
\toprule
\multirow{2}{*}{\textbf{Category}}   & \multirow{2}{*}{\textbf{Task}} & \multicolumn{6}{c|}{\textbf{Closed-Source Models}}                                                                                                               & \multicolumn{4}{c|}{\textbf{Open-Source Models}}                                                                                                                                                                                             & \multicolumn{2}{c}{\textbf{Prompt Enhancement}} \\ \cmidrule(l){3-14} 
                            &                       & Veo3.1 & Veo2  & Sora2 & \begin{tabular}[c]{@{}c@{}}Pixel\\ Verse-V5\end{tabular} & \begin{tabular}[c]{@{}c@{}}Seedance\\ 1.0-pro\end{tabular} & Wan2.5 & \begin{tabular}[c]{@{}c@{}}CogVideoX\\ 1.5 5B\end{tabular} & \begin{tabular}[c]{@{}c@{}}Hunyuan\\ Video\end{tabular} & \begin{tabular}[c]{@{}c@{}}Wan2.1\\ 14B\end{tabular} & \begin{tabular}[c]{@{}c@{}}Wan2.2\\ A14B\end{tabular} & Veo3.1 PE          & Sora2 PE          \\ \midrule
\multirow{8}{*}{Humanity}   & transportation        & 80.00  & 54.50 & 71.56 & 73.44                                                    & 68.19                                                      & 67.53  & 37.15                                                      & 48.72                                                   & 53.24                                                & 50.67                                                 & 78.50              & 69.38             \\
                            & sport                 & 78.08  & 62.80 & 73.48 & 71.66                                                    & 65.87                                                      & 74.95  & 44.98                                                      & 58.74                                                   & 56.06                                                & 59.66                                                 & 82.36              & 79.65             \\
                            & social                & 74.88  & 51.09 & 77.77 & 65.23                                                    & 64.24                                                      & 61.18  & 36.54                                                      & 49.49                                                   & 51.42                                                & 51.72                                                 & 75.08              & 75.22             \\
                            & safety                & 82.79  & 48.18 & 80.13 & 75.13                                                    & 65.79                                                      & 57.40  & 45.46                                                      & 47.79                                                   & 52.93                                                & 59.36                                                 & 85.49              & 87.88             \\
                            & festival              & 80.83  & 60.43 & 84.35 & 80.03                                                    & 78.65                                                      & 61.54  & 48.28                                                      & 61.81                                                   & 61.71                                                & 68.03                                                 & 90.03              & 89.74             \\
                            & dress                 & 90.87  & 87.18 & 89.82 & 86.86                                                    & 88.02                                                      & 85.10  & 72.76                                                      & 81.97                                                   & 81.25                                                & 85.74                                                 & 94.15              & 93.67             \\
                            & food                  & 90.94  & 78.25 & 88.86 & 78.65                                                    & 72.90                                                      & 67.67  & 57.68                                                      & 66.03                                                   & 65.61                                                & 60.97                                                 & 93.36              & 85.68             \\
                            & emotion               & 69.21  & 56.67 & 63.60 & 66.34                                                    & 59.85                                                      & 45.46  & 33.68                                                      & 42.89                                                   & 52.07                                                & 40.12                                                 & 59.54              & 55.57             \\ \midrule
\multirow{7}{*}{Science}    & chemistry             & 81.13  & 71.20 & 85.78 & 75.21                                                    & 66.61                                                      & 65.73  & 45.90                                                      & 36.87                                                   & 46.25                                                & 51.62                                                 & 86.59              & 81.63             \\
                            & physics               & 79.09  & 56.99 & 76.41 & 72.32                                                    & 73.44                                                      & 76.34  & 34.78                                                      & 36.18                                                   & 54.30                                                & 54.17                                                 & 74.49              & 82.25             \\
                            & biology               & 83.31  & 52.48 & 84.14 & 73.63                                                    & 75.68                                                      & 74.97  & 47.30                                                      & 40.95                                                   & 55.81                                                & 52.44                                                 & 81.15              & 76.38             \\
                            & earth                 & 57.27  & 35.46 & 52.92 & 52.02                                                    & 47.81                                                      & 49.12  & 39.42                                                      & 39.12                                                   & 50.27                                                & 48.94                                                 & 64.35              & 51.12             \\
                            & math                  & 61.52  & 55.34 & 70.09 & 57.86                                                    & 57.31                                                      & 65.76  & 32.78                                                      & 39.48                                                   & 42.68                                                & 43.33                                                 & 61.28              & 72.92             \\
                            & medicine              & 63.38  & 53.53 & 64.64 & 64.18                                                    & 61.11                                                      & 56.09  & 36.18                                                      & 44.91                                                   & 46.72                                                & 47.84                                                 & 68.14              & 80.62             \\
                            & life                  & 74.51  & 51.50 & 65.86 & 67.54                                                    & 65.28                                                      & 55.44  & 42.77                                                      & 37.21                                                   & 52.43                                                & 55.62                                                 & 83.68              & 73.50             \\ \midrule
\multirow{10}{*}{Game}      & chess                 & 20.94  & 16.25 & 45.94 & 28.12                                                    & 36.84                                                      & 28.91  & 28.28                                                      & 31.09                                                   & 21.88                                                & 33.28                                                 & 23.44              & 50.78             \\
                            & puzzle                & 64.43  & 41.56 & 44.95 & 58.39                                                    & 52.63                                                      & 59.90  & 27.03                                                      & 24.06                                                   & 38.65                                                & 36.67                                                 & 53.54              & 49.95             \\
                            & gomoku                & 39.43  & 41.67 & 56.98 & 41.20                                                    & 51.72                                                      & 46.51  & 46.61                                                      & 32.24                                                   & 36.98                                                & 44.90                                                 & 46.82              & 71.72             \\
                            & sudoku                & 52.36  & 30.69 & 71.81 & 64.31                                                    & 51.76                                                      & 51.11  & 46.81                                                      & 34.72                                                   & 43.33                                                & 53.61                                                 & 51.67              & 67.50             \\
                            & maze                  & 56.60  & 39.17 & 54.44 & 45.28                                                    & 52.64                                                      & 57.57  & 28.89                                                      & 26.67                                                   & 38.89                                                & 44.51                                                 & 59.31              & 51.25             \\
                            & minesweeper           & 70.00  & 42.50 & 71.41 & 68.75                                                    & 68.44                                                      & 73.44  & 47.66                                                      & 47.81                                                   & 50.78                                                & 50.47                                                 & 57.66              & 78.12             \\
                            & number puzzle         & 50.83  & 30.83 & 46.11 & 32.50                                                    & 41.67                                                      & 47.78  & 26.39                                                      & 22.22                                                   & 33.06                                                & 46.11                                                 & 32.22              & 52.22             \\
                            & sticks                & 39.53  & 21.98 & 54.17 & 54.01                                                    & 43.54                                                      & 55.31  & 38.96                                                      & 23.44                                                   & 33.39                                                & 47.55                                                 & 40.99              & 57.24             \\
                            & xiangqi               & 34.27  & 19.41 & 47.64 & 39.09                                                    & 44.08                                                      & 32.95  & 51.37                                                      & 25.82                                                   & 27.47                                                & 34.10                                                 & 31.36              & 48.68             \\
                            & Go                    & 41.44  & 38.10 & 58.26 & 43.53                                                    & 46.15                                                      & 49.40  & 44.42                                                      & 33.11                                                   & 22.20                                                & 51.64                                                 & 32.74              & 59.30             \\ \midrule
\multirow{2}{*}{Hypothesis} & subjective            & 74.06  & 46.08 & 62.73 & 58.87                                                    & 49.91                                                      & 49.42  & 35.56                                                      & 38.50                                                   & 46.24                                                & 44.90                                                 & 75.36              & 73.02             \\
                            & objective             & 81.31  & 61.21 & 73.27 & 86.12                                                    & 68.66                                                      & 65.64  & 46.91                                                      & 50.52                                                   & 58.02                                                & 61.72                                                 & 81.91              & 84.32             \\ \midrule
\multirow{3}{*}{Semantics}  & idioms                & 76.53  & 60.89 & 72.14 & 72.11                                                    & 65.56                                                      & 60.00  & 51.13                                                      & 52.21                                                   & 59.05                                                & 57.83                                                 & 83.82              & 75.13             \\
                            & metaphors             & 81.36  & 78.75 & 70.37 & 80.56                                                    & 74.04                                                      & 72.12  & 67.98                                                      & 68.93                                                   & 71.52                                                & 70.54                                                 & 79.73              & 82.07             \\
                            & definition            & 87.25  & 71.50 & 80.08 & 76.38                                                    & 76.15                                                      & 82.61  & 58.10                                                      & 55.78                                                   & 58.68                                                & 64.01                                                 & 86.62              & 79.01             \\ \midrule
\multirow{10}{*}{Vision}    & anomaly               & 42.19  & 34.69 & 32.81 & 33.12                                                    & 37.19                                                      & 36.25  & 8.12                                                       & 24.06                                                   & 21.56                                                & 28.44                                                 & 38.44              & 35.62             \\
                            & color                 & 63.89  & 26.39 & 22.92 & 36.81                                                    & 51.39                                                      & 54.17  & 22.22                                                      & 29.17                                                   & 30.56                                                & 33.33                                                 & 62.50              & 38.19             \\
                            & count                 & 20.14  & 18.75 & 34.03 & 26.56                                                    & 38.89                                                      & 42.36  & 8.33                                                       & 23.61                                                   & 22.22                                                & 38.89                                                 & 26.39              & 38.19             \\
                            & direction             & 32.50  & 30.00 & 30.00 & 35.00                                                    & 48.75                                                      & 32.81  & 16.88                                                      & 17.50                                                   & 30.00                                                & 44.38                                                 & 36.88              & 28.75             \\
                            & position              & 37.50  & 43.12 & 41.88 & 44.38                                                    & 37.50                                                      & 45.62  & 21.88                                                      & 28.13                                                   & 32.50                                                & 36.88                                                 & 28.12              & 42.50             \\
                            & shape                 & 46.88  & 27.34 & 39.58 & 47.22                                                    & 54.17                                                      & 36.11  & 18.06                                                      & 27.78                                                   & 31.94                                                & 37.50                                                 & 50.78              & 40.28             \\
                            & size                  & 47.92  & 30.73 & 38.02 & 43.75                                                    & 43.75                                                      & 51.14  & 20.83                                                      & 42.71                                                   & 36.98                                                & 41.67                                                 & 46.35              & 31.25             \\
                            & style                 & 47.50  & 23.12 & 43.75 & 36.88                                                    & 31.25                                                      & 43.75  & 7.50                                                       & 20.62                                                   & 25.62                                                & 32.50                                                 & 47.50              & 46.88             \\
                            & view                  & 51.25  & 29.38 & 31.25 & 50.62                                                    & 45.62                                                      & 53.75  & 19.38                                                      & 42.50                                                   & 31.88                                                & 38.12                                                 & 51.25              & 32.50             \\
                            & motion                & 55.31  & 51.04 & 41.77 & 56.88                                                    & 56.88                                                      & 53.12  & 27.29                                                      & 31.25                                                   & 42.71                                                & 50.21                                                 & 57.92              & 53.65             \\ \bottomrule
\end{tabular}
\end{table*}

%% file: supp_secs/open_figure.tex
\begin{figure*}[t]
  \centerline{\includegraphics[width=\textwidth]{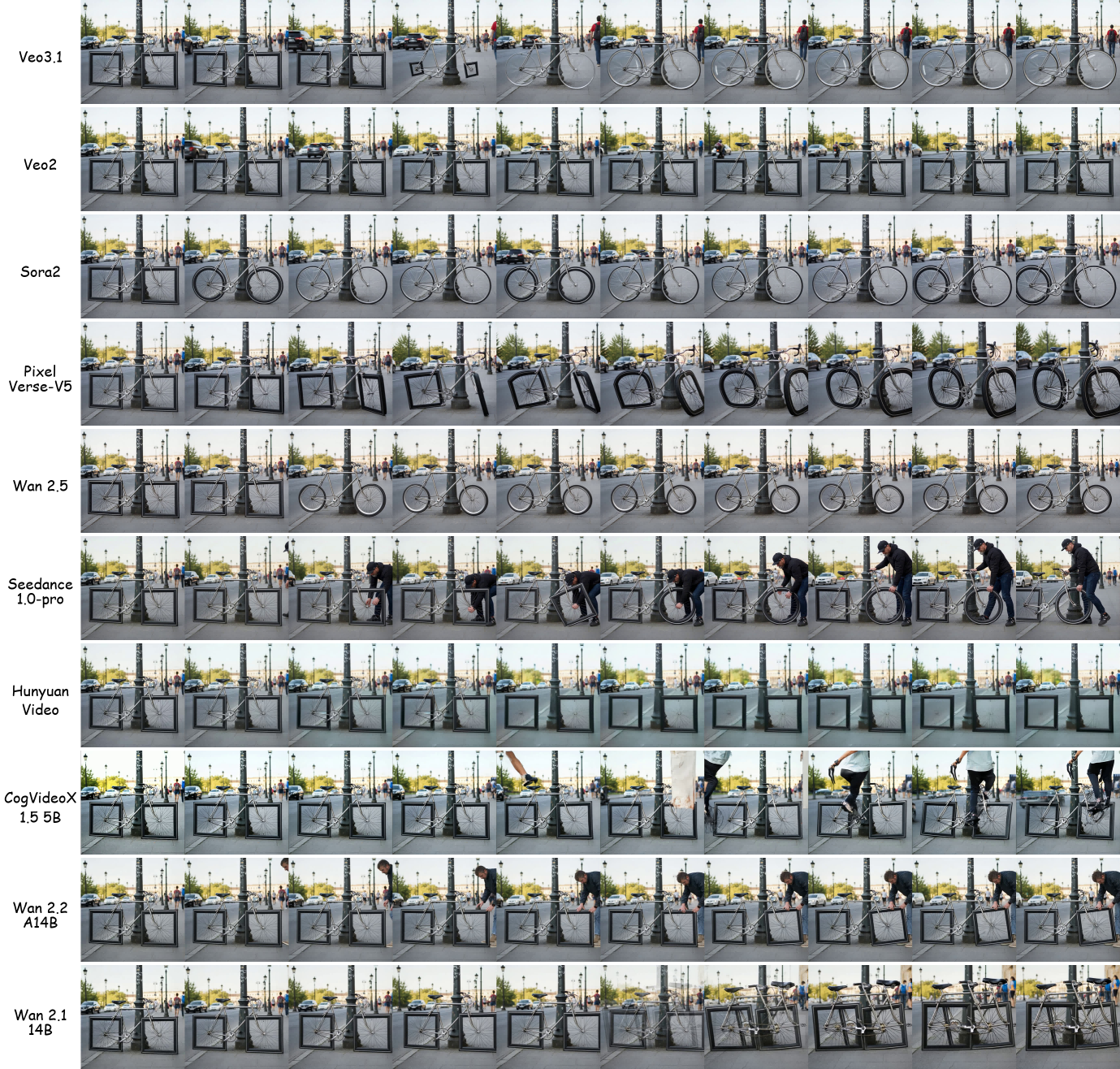}} 
  \caption{Qualitative results of the Anomaly task across 10 video models.
  \textbf{\textit{Prompt}}: Please correct the bicycle's wheel shape anomaly so that the vehicle can function properly.
  }
  \label{supp_fig:anomaly}
\end{figure*}

\begin{figure*}[t]
  \centerline{\includegraphics[width=\textwidth]{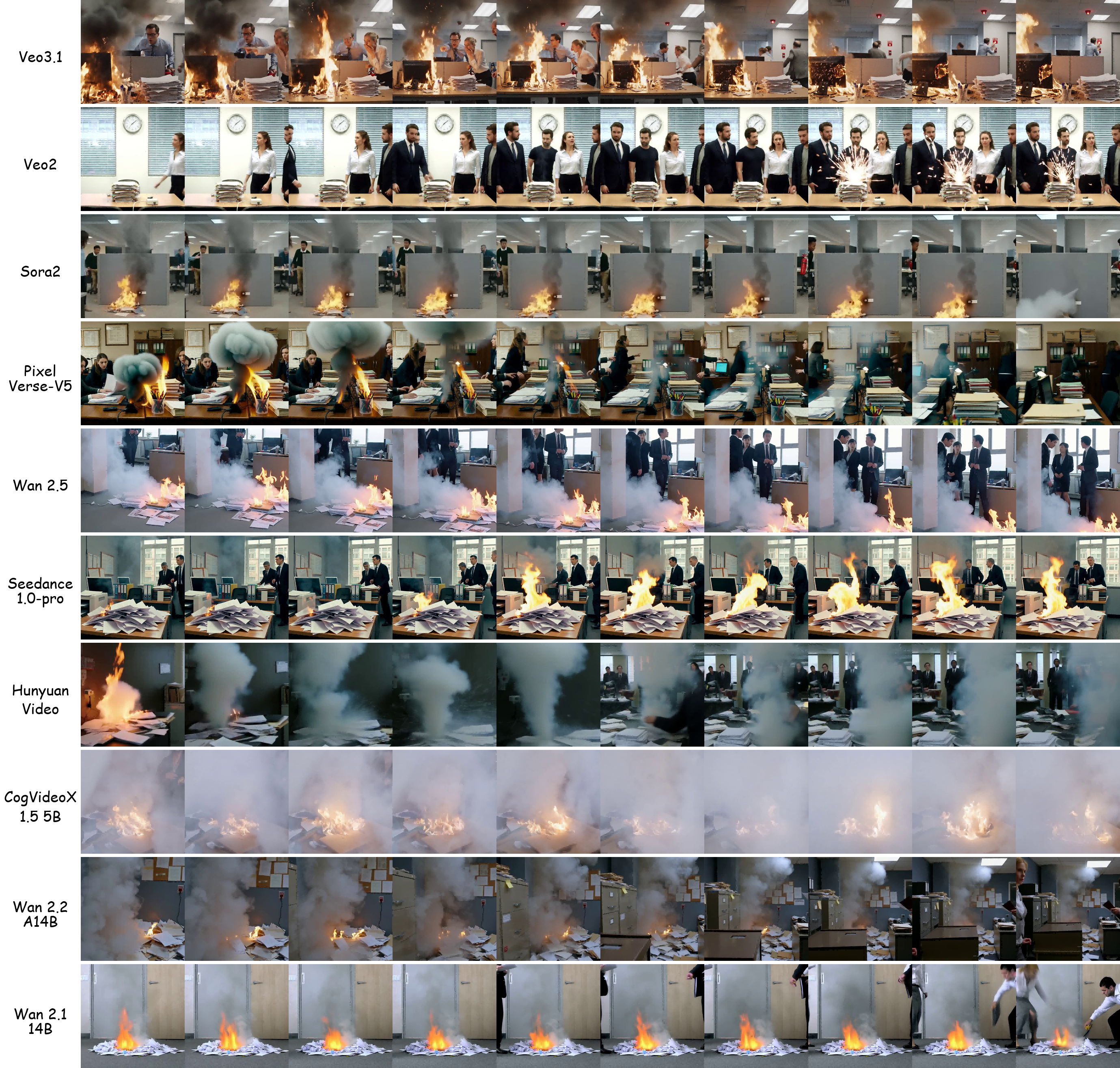}} 
  \caption{Qualitative results of the Safety task across 10 video models.
  \textbf{\textit{Prompt}}: In an office, thick smoke and small flames are rising from a pile of paper near a power socket. Several employees notice the fire spreading quickly across nearby papers and office supplies. They look around the room, searching for a way to extinguish it.
  }
  \label{supp_fig:safety}
\end{figure*}

\begin{figure*}[t]
  \centerline{\includegraphics[width=\textwidth]{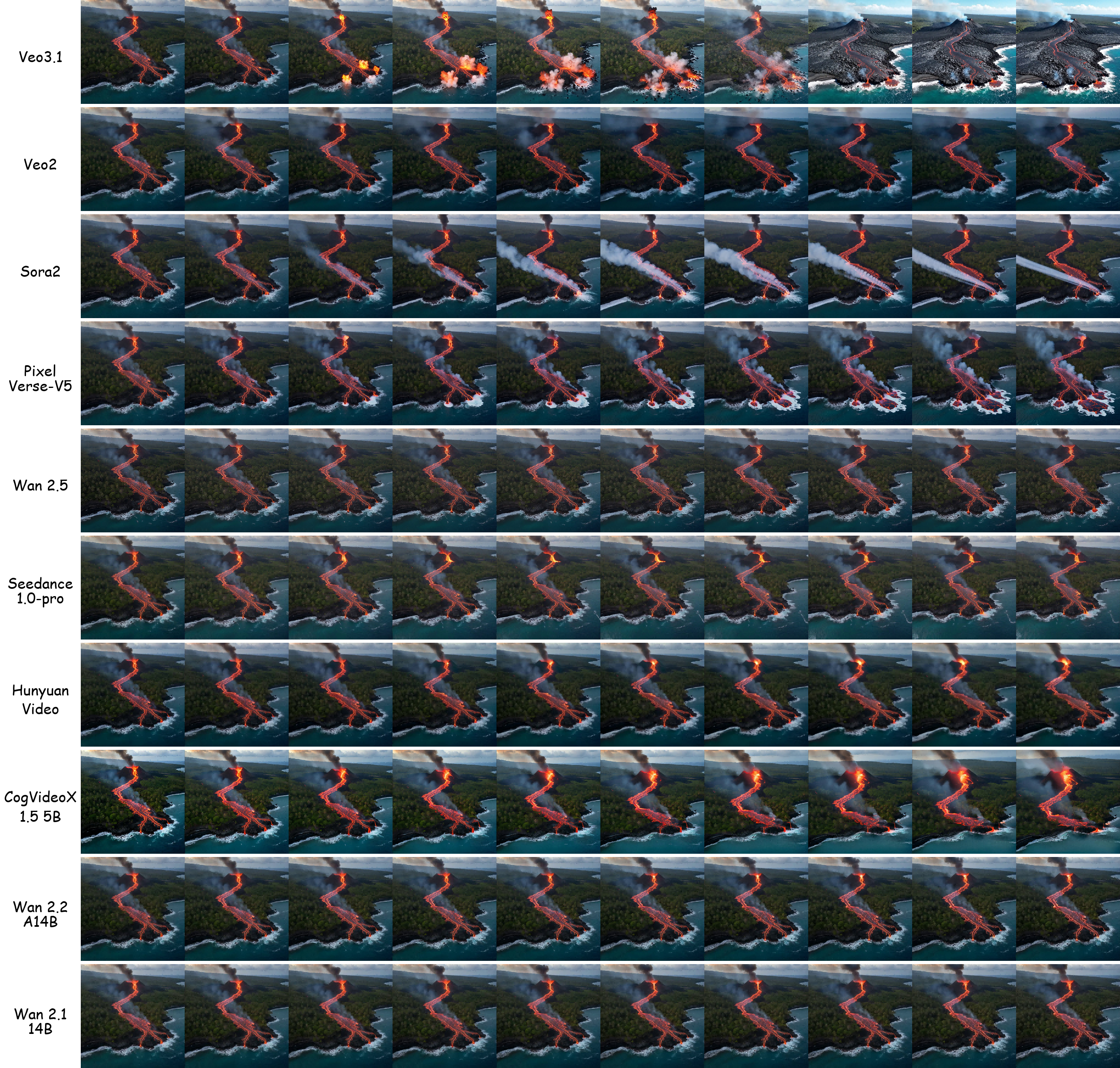}} 
  \caption{Qualitative results of the Earth task across 10 video models.
  \textbf{\textit{Prompt}}: Observe the lava flow over several years under continuous volcanic activity, noting changes along its path.
  }
  \label{supp_fig:earth}
\end{figure*}

\begin{figure*}[t]
  \centerline{\includegraphics[width=\textwidth]{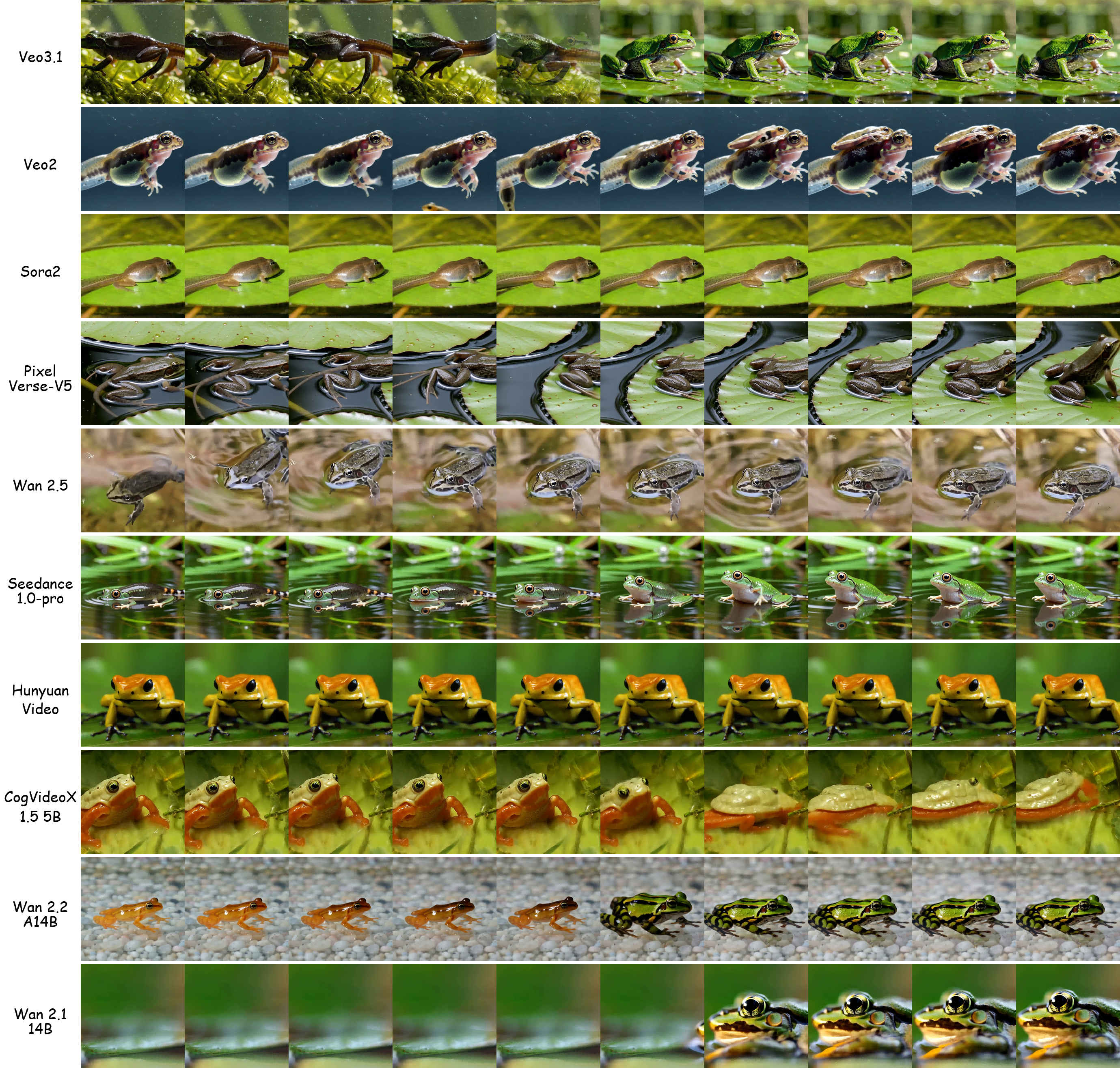}} 
  \caption{Qualitative results of the Life task across 10 video models.
  \textbf{\textit{Prompt}}: The process of a tadpole transforming into a frog.
  }
  \label{supp_fig:life}
\end{figure*}

\begin{figure*}[t]
  \centerline{\includegraphics[width=\textwidth]{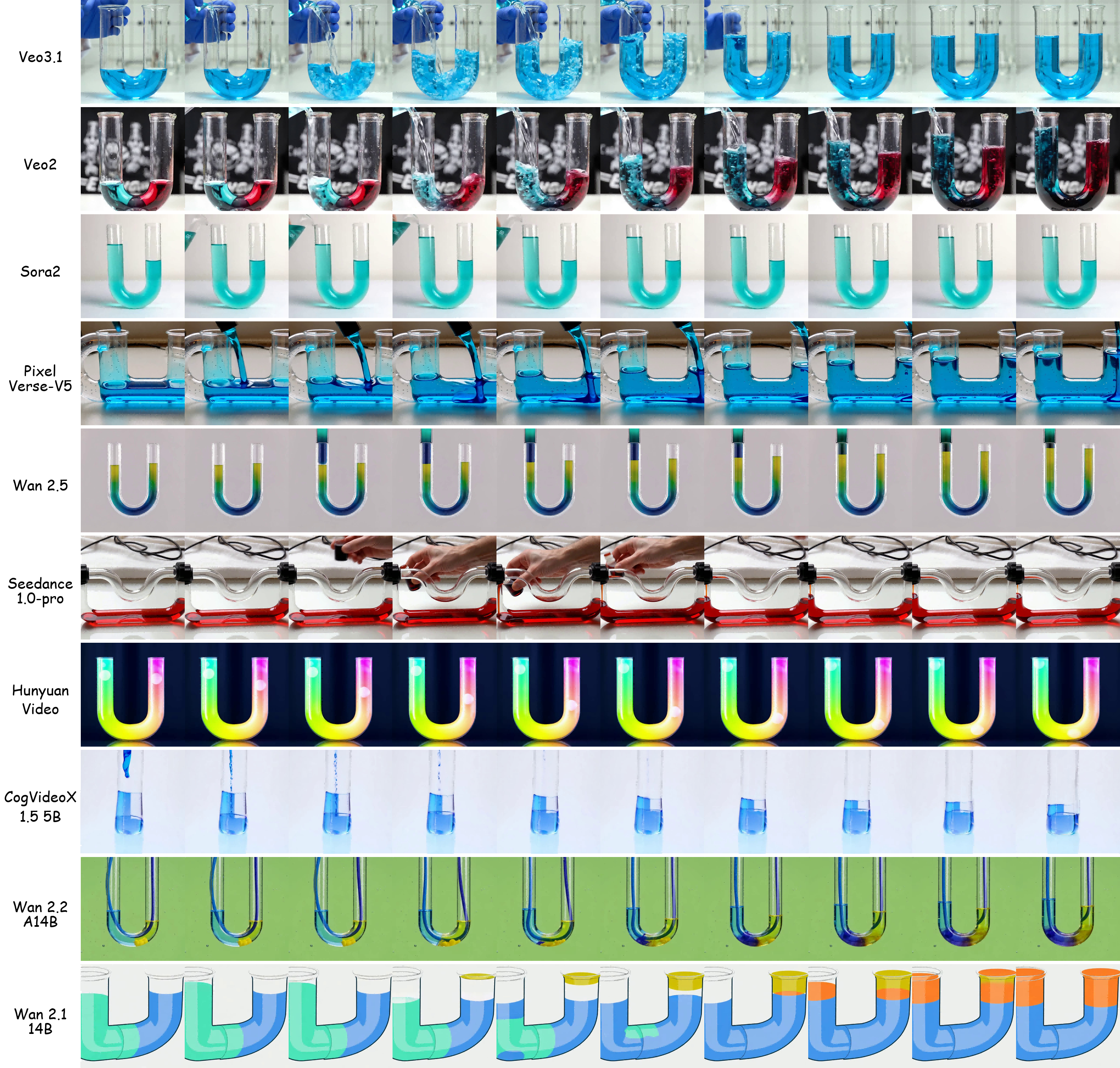}} 
  \caption{Qualitative results of the Physics task across 10 video models.
  \textbf{\textit{Prompt}}: Fill a U-shaped transparent tube with colored water so that both arms have the same initial height. Add extra colored water to the left arm and observe the water levels in both arms.
  }
  \label{supp_fig:physics}
\end{figure*}

\begin{figure*}[t]
  \centerline{\includegraphics[width=\textwidth]{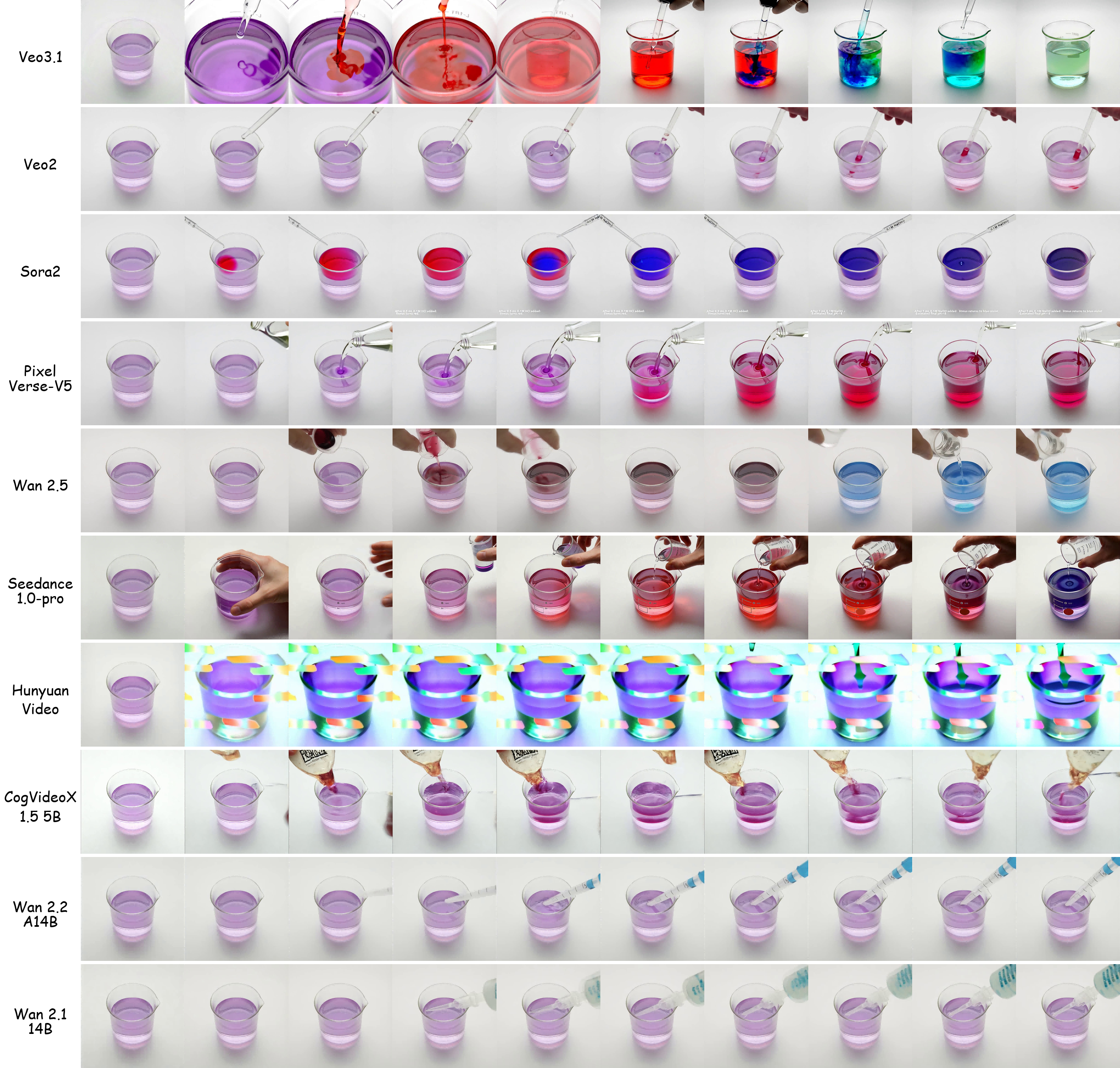}} 
  \caption{Qualitative results of the Chemistry task across 10 video models.
  \textbf{\textit{Prompt}}: First, add 0.5 mL of 0.1 M dilute hydrochloric acid and observe the litmus color change; then add 1 mL of 0.1 M dilute sodium hydroxide, recording the color change.
  }
  \label{supp_fig:chemistry}
\end{figure*}

\begin{figure*}[t]
  \centerline{\includegraphics[width=\textwidth]{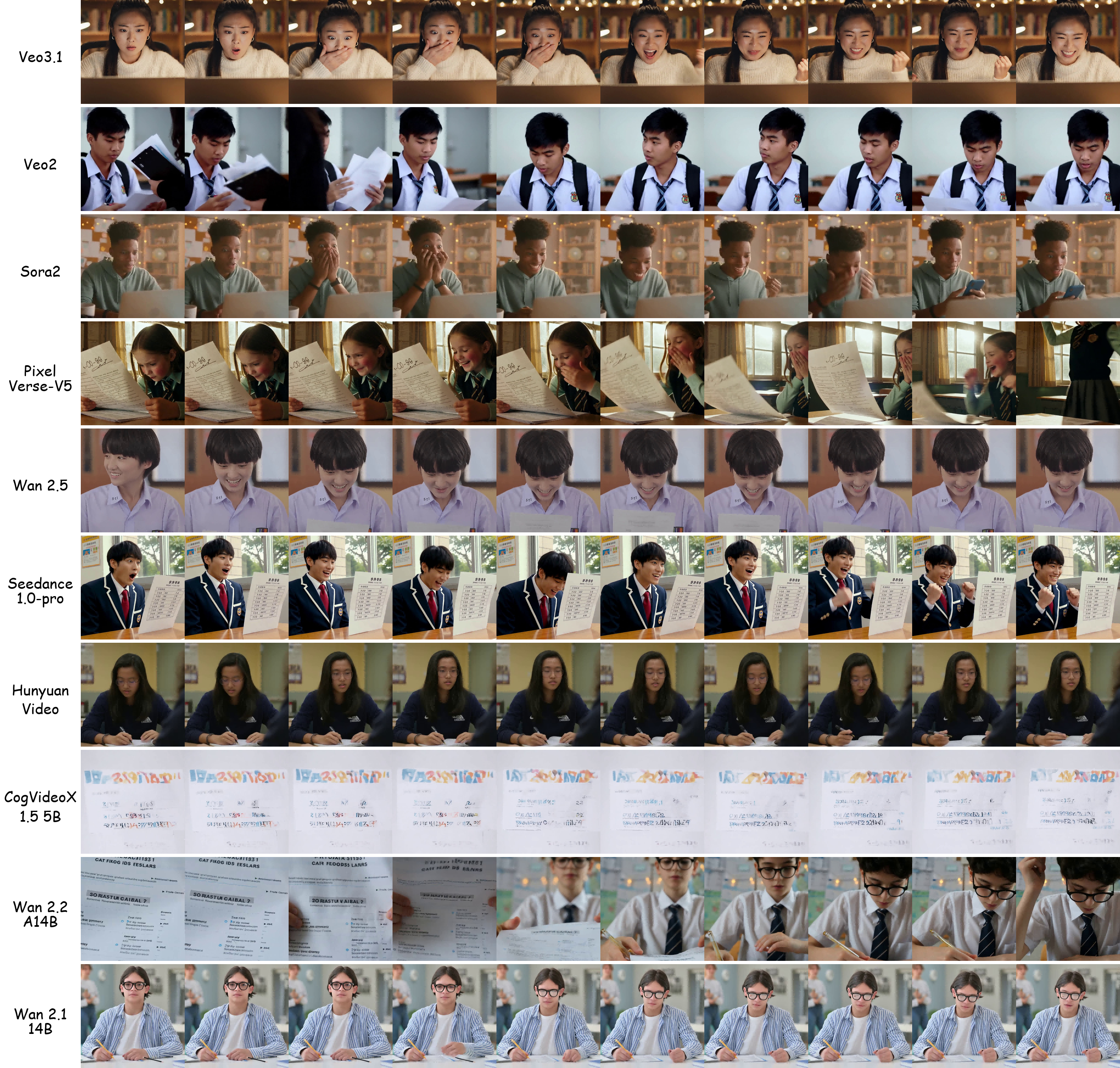}} 
  \caption{Qualitative results of the Emotion task across 10 video models.
  \textbf{\textit{Prompt}}: A student has just learned that their exam results are excellent, exceeding expectations.
  }
  \label{supp_fig:emotion}
\end{figure*}

\begin{figure*}[t]
  \centerline{\includegraphics[width=\textwidth]{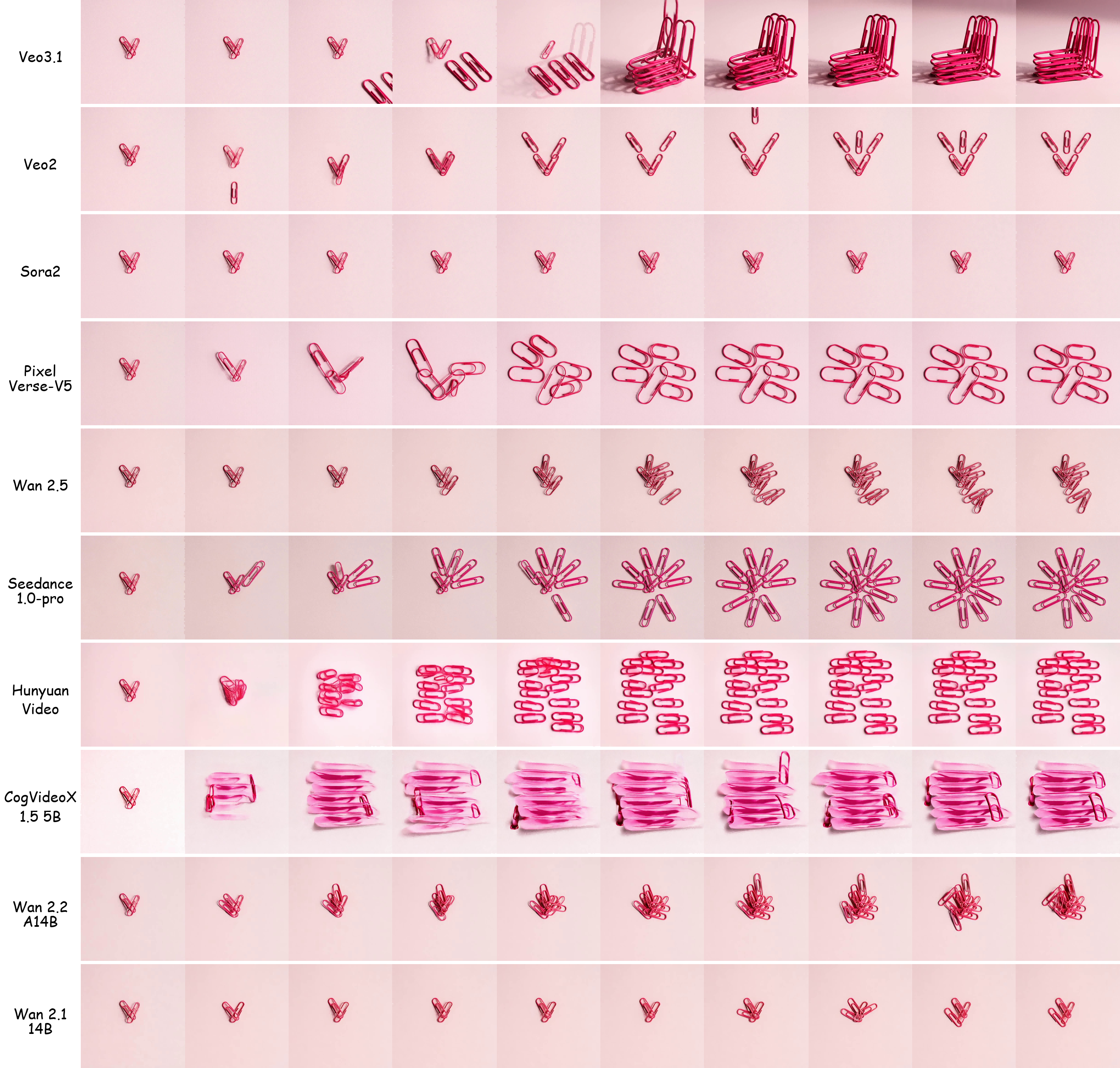}} 
  \caption{Qualitative results of the Count task across 10 video models.
  \textbf{\textit{Prompt}}: Increase the number of pink paperclips in the input image so that there are exactly five partially overlapping paperclips, arranged in a visually natural stacked formation where each clip slightly overlaps the next.
  }
  \label{supp_fig:color}
\end{figure*}

%% file: supp_secs/open_prompt.tex
\begin{figure*}
\centering
\begin{promptbox}{System Prompt for Evaluation}

You are a video quality assessment expert. Respond using the following format exactly:
\vspace{0.5em}

\texttt{<think></think>}

\texttt{<answer></answer>}
\vspace{0.5em}

In \texttt{<think></think>}, first provide a concise, concrete description of the task. 
Next, give an objective understanding/summary of the input video (what happens in the video, salient visual / content elements). 
For each checklist question, give a brief factual justification (one or two short sentences) explaining why you rated that question as you did. 
Do not reveal private chain-of-thought; provide only a concise, professional justification for each answer.
\vspace{0.5em}

In \texttt{<answer></answer>}, write the answers to all checklist questions as a JSON-style list of labels in order, for example:
\texttt{["Good", "Medium", "Poor", ...]}  
The list length must exactly match the number of checklist questions.  
Each item must be one of \textbf{``Good''}, \textbf{``Medium''}, or \textbf{``Poor''}.  
Do not include any other text, commentary, or explanation in \texttt{<answer>}.
\vspace{0.5em}
You will be provided:
\begin{itemize}
  \item a video
  \item the generation prompt
  \item an optional implicit explanation
  \item an optional input image
  \item an optional ground-truth image
  \item a set of checklist questions
\end{itemize}
\vspace{0.5em}
\textbf{Definitions}
\vspace{0.5em}
\begin{itemize}
  \item \textbf{generation prompt}: the single textual instruction given to the video-generation model.
  \item \textbf{implicit explanation}: an optional detailed explanation of the intended outcome.
  \item \textbf{input image}: optional visual input provided to the model.
  \item \textbf{ground truth image}: the expected final frame or expected motion trajectory.
  \item \textbf{checklist questions}: specific quality or content questions to assess.
\end{itemize}
\vspace{0.5em}
\textbf{Notes}
\vspace{0.5em}
\begin{itemize}
  \item \textbf{``Good''} $\rightarrow$ The video clearly and fully satisfies the question.
  \item \textbf{``Medium''} $\rightarrow$ The video partially satisfies the question or has minor issues.
  \item \textbf{``Poor''} $\rightarrow$ The video fails to satisfy the question or has major issues.
\end{itemize}
\vspace{0.5em}
\textbf{Rules}
\vspace{0.5em}
\begin{enumerate}
  \item If the premise of a question is not met by the video (e.g., the question asks about an element that never appears), rate that question \textbf{``Poor''}.
  \item Evaluate each question independently; do not let other answers influence a question's rating.
  \item Provide concise justifications in \texttt{<think>} and only the ordered list of labels in \texttt{<answer>}.
  \item The response must \textbf{strictly follow} the \texttt{<think></think><answer></answer>} format and include nothing outside those tags.
\end{enumerate}
\end{promptbox}
\caption{System prompt template for the evaluation of generated videos.}
\label{supp_fig:eval_prompt}
\end{figure*}

\begin{figure*}
\centering
\begin{promptbox}{System Prompt for Image-to-scenario Pipeline}
For each sample, you must output two keys: \texttt{prompt} and \texttt{implicit\_explanation}, based on the provided image.
Follow the detailed rules below precisely to ensure clarity and consistency.

\vspace{0.5em}
\begin{enumerate}
\item \textbf{Prompt:}
\vspace{0.5em}
\begin{itemize}
\item Describe the \textbf{action or external event} applied to the provided image — the only textual input to the model.
\item \textbf{Requirements:}
\begin{itemize}
  \item Begin with a single sentence briefly describing the visual content of the \textit{initial\_scenario} (e.g., ``In the image, the beaker on the left contains hydrochloric acid, and the beaker on the right contains a sodium hydroxide solution.'')
  \item Must contain \textbf{exactly one main action or event.}
  \item Exclude results, explanations, or causes.
  \item Keep it \textbf{precise} and maintain the same style and perspective as the initial scene.
  \item Use unambiguous, temporally clear language.
  \item Must not leak any information from \texttt{implicit\_explanation.}
  \item Ensure \textbf{standalone completeness}: the description must remain logically self-contained even when no initial scenario is provided.
  \item Clearly specify all references in each sentence; avoid vague terms such as ``some'', ``certain'', ``a kind of'', or ``unknown'' — always use explicit, concrete nouns.
\end{itemize}
\end{itemize}

\vspace{0.5em}
\item \textbf{Implicit Explanation:}
\vspace{0.5em}
\begin{itemize}
\item Explain the hidden reasoning behind the prompt: the mechanism that will unfold after the prompt’s action.
\item \textbf{Requirements:}
\begin{itemize}
  \item Describe factual mechanisms and expected consequences logically.
  \item Include a clear \textbf{cause → process → result} sequence.
  \item Use precise, domain-correct terminology (no metaphors or emotional tone).
  \item Maintain full consistency with both \textit{initial\_scenario} and \textit{prompt.}
\end{itemize}
\end{itemize}
\end{enumerate}

\vspace{0.5em}
\noindent\textbf{Cautions:}
\vspace{0.5em}
\begin{itemize}
\item All samples must maintain \textbf{high-school level reasoning difficulty}: complex enough to require logical understanding but not specialized technical expertise.
\item Scenarios must remain within the \textbf{real-world physical scale}: avoid microscopic (e.g., atomic, cellular) and macroscopic (e.g., planetary, galactic) contexts.
\item Ensure \textbf{internal logical coherence}: objects, actions, and outcomes must follow physically plausible relationships.
\item All components (\textit{initial\_scenario}, \textit{prompt}, \textit{implicit\_explanation}) must align stylistically and factually.
\item Avoid fantasy or science-fiction elements unless explicitly grounded in real-world analogs.
\item Maintain visual realism and a neutral tone — no symbolic, emotional, or metaphorical descriptions.
\item Each generated sample should \textbf{strictly correspond to the specific task requirements and concepts}, with no ambiguity in objects, actions, or reasoning related to the task.
\end{itemize}
\end{promptbox}
\caption{System prompt template for Image-to-scenario pipeline designed for I2V tasks.}
\label{supp_fig:I2S}
\end{figure*}

\begin{figure*}
\begin{promptbox}{System Prompt for Scenario-to-image Pipeline}
For each sample, you must output \textbf{3 keys}: \texttt{initial\_scenario}, \texttt{prompt}, and \texttt{implicit\_explanation}.  
Follow the detailed rules below precisely to ensure clarity and consistency.

\begin{enumerate}
\vspace{0.5em}
\item \textbf{initial\_scenario:}
\vspace{0.5em}
  \begin{itemize}
    \item Describe the static starting state of the scene in precise, visual, and objective language. This defines what the model initially ``sees'' before any change occurs.
    \item \textbf{Requirements:}
    \begin{itemize}
      \item The description should be detailed and concise, including visual objects, spatial arrangement, lighting, composition, and style, etc.
      \item No actions, no temporal hints, no implied future events.
      \item Should logically support the descriptions in \texttt{prompt}.
      \item Do not include any visible text, letters, characters, or numbers in the scene description (e.g., avoid ``a book with the title `History' on the cover'' or ``a clock showing the number 12'').
      \item Clearly specify all references in each sentence; avoid vague terms such as ``some'', ``certain'', or ``unknown'' --- always use explicit, concrete nouns.
    \end{itemize}
  \end{itemize}

\vspace{0.5em}
\item \textbf{prompt:}
\vspace{0.5em}
  \begin{itemize}
    \item Describe the action or external event applied to the initial scene — the only textual input to the generative model.
    \item \textbf{Requirements:}
    \begin{itemize}
      \item Begin with a single sentence briefly describing the visual content of the \texttt{initial\_scenario} (e.g., ``In the image, the beaker on the left contains hydrochloric acid, and the beaker on the right contains a sodium hydroxide solution.'')
      \item Must contain exactly one main action or event.
      \item No results, explanations, or causes.
      \item Maintain the same style and perspective as the initial scene.
      \item Must not leak any information from \texttt{implicit\_explanation}.
      \item Standalone completeness: The description must be logically self-contained even when no initial scenario is provided.
      \item Ensure continuity between actions and the initial scene.
      \item Clearly specify all references in each sentence; avoid vague terms such as ``some'', ``certain'', ``a kind of'', ``unknown'' — always use explicit, concrete nouns.
    \end{itemize}
  \end{itemize}

\vspace{0.5em}
\item \textbf{implicit\_explanation:}
\vspace{0.5em}
  \begin{itemize}
    \item Explain the hidden reasoning behind the prompt: the mechanism that will unfold after the prompt’s action.
    \item \textbf{Requirements:}
    \begin{itemize}
      \item Describe factual mechanisms and expected consequences logically.
      \item Include cause $\rightarrow$ process $\rightarrow$ result sequence.
      \item Use precise, domain-correct terminology (no metaphors or emotional tone).
      \item Must stay fully consistent with \texttt{initial\_scenario} and \texttt{prompt}.
    \end{itemize}
  \end{itemize}
\end{enumerate}

\vspace{0.5em}
\noindent\textbf{Cautions}
\vspace{0.5em}
\begin{itemize}
  \item Scenarios must remain strictly within the real-world physical scale: avoid microscopic (e.g., atomic, cellular) and macroscopic (e.g., planetary, galactic) contexts.
  \item Ensure internal logical coherence: objects, actions, and outcomes must follow physically plausible relationships.
  \item All components (\texttt{initial\_scenario}, \texttt{prompt}, \texttt{implicit\_explanation}) must align stylistically and factually with one another.
  \item Each generated sample should strictly correspond to the specific task requirements and concepts, with no ambiguity in objects, actions, or reasoning related to the task.
\end{itemize}
\end{promptbox}
\caption{System prompt template for Scenario-to-image pipeline. \textit{Initial scenarios} are instructed to T2I systems to generate images.}
\label{supp_fig:S2I}
\end{figure*}

\begin{figure*}
\begin{promptbox}{System Prompt for Checklist Generation}
You are an expert evaluator specialized in assessing video generation models.  
Your task is to create a \textbf{checklist} composed of questions that will be used by a multimodal model to judge the quality of a generated video.

\vspace{0.5em}
\textbf{Inputs}
\vspace{0.5em}
\begin{itemize}
  \item \textbf{prompt}: the only textual instruction used by the video generation model.
  \item \textbf{input image or video (optional)}: the only visual input provided to the model.
  \item \textbf{gt\_image (optional)}: the expected final frame or the expected motion trajectory.
  \item \textbf{implicit\_explanation (optional)}: a detailed explanation of the intended outcome.
\end{itemize}

\vspace{0.5em}
\textbf{Evaluation Dimensions}
\vspace{0.5em}
\begin{itemize}
  \item \textbf{Instruction Following} – Whether the video correctly follows the prompt’s instruction.
  \item \textbf{Visual Consistency} – Whether objects, actors, and camera motions remain coherent across space and time — including alignment, motion realism, and smooth scene transitions.
  \item \textbf{Rule Coherence} – Whether the video logically reflects causal, physical, or rule-based reasoning implied by the prompt.
  \item \textbf{Visual Fidelity} – Whether the video maintains high visual quality — including clarity, color stability, and absence of artifacts or distortions.
\end{itemize}

\vspace{0.5em}
\textbf{Checklist Design Rules}
\vspace{0.5em}
\begin{enumerate}
  \item Include \textbf{at least five questions}.
  \item Each question must refer to \textbf{specific objects, actions, or outcomes} described in the inputs.
  \item Include \textbf{at least one question} that explicitly tests the model’s \textbf{reasoning ability}.
  \item Each reasoning-related question must focus on only one reasoning aspect (e.g., causal reasoning, physical reasoning, spatial reasoning) — do not combine multiple reasoning types in a single question.
  \item Phrase every question so that answering \textbf{“Yes”} means the generated video meets the expected criteria, and \textbf{“No”} means it does not.
  \item Use only information that appears in the \texttt{prompt}, \texttt{input image / video}, \texttt{gt\_image}, and \texttt{explicit\_explanation}.
  \item If part of the \texttt{explicit\_explanation} describes something that is not a necessary result of the prompt, you may ignore it.
  \item If multiple reasoning aspects are involved, design separate questions for each reasoning aspect.
  \item The final checklist must include at least one question corresponding to each of the four dimensions: \textbf{Instruction Following}, \textbf{Visual Consistency}, \textbf{Rule Coherence}, and \textbf{Visual Fidelity}.
\end{enumerate}

\vspace{0.5em}
\textbf{Output Format}  
\vspace{0.5em}

Return the checklist as a valid Python list of JSON objects, where each JSON object uses the dimension name as the key and the corresponding Yes/No question as the value.
\end{promptbox}
\caption{System prompt template for checklist generation across 4 evaluation metrics. }
\label{supp_fig:checklist}
\end{figure*}

%% file: main.bbl
\begin{thebibliography}{95}
\providecommand{\natexlab}[1]{#1}
\providecommand{\url}[1]{\texttt{#1}}
\expandafter\ifx\csname urlstyle\endcsname\relax
  \providecommand{\doi}[1]{doi: #1}\else
  \providecommand{\doi}{doi: \begingroup \urlstyle{rm}\Url}\fi

\bibitem[Agarwal et~al.(2025)Agarwal, Ali, Bala, Balaji, Barker, Cai, Chattopadhyay, Chen, Cui, Ding, et~al.]{agarwal2025cosmos}
Niket Agarwal, Arslan Ali, Maciej Bala, Yogesh Balaji, Erik Barker, Tiffany Cai, Prithvijit Chattopadhyay, Yongxin Chen, Yin Cui, Yifan Ding, et~al.
\newblock Cosmos world foundation model platform for physical ai.
\newblock \emph{arXiv preprint arXiv:2501.03575}, 2025.

\bibitem[{Alibaba Cloud (Wan Series)}(2025)]{wan2.5-model2025}
{Alibaba Cloud (Wan Series)}.
\newblock Wan 2.5, 2025.

\bibitem[{Anthropic}(2025{\natexlab{a}})]{claudehaiku45}
{Anthropic}.
\newblock Introducing claude haiku 4.5, 2025{\natexlab{a}}.

\bibitem[{Anthropic}(2025{\natexlab{b}})]{claudesonnet45}
{Anthropic}.
\newblock System card: Claude sonnet 4.5.
\newblock Technical report, Anthropic, 2025{\natexlab{b}}.
\newblock Technical report, Anthropic, September 2025.

\bibitem[Bai et~al.(2025)Bai, Chen, Liu, Wang, Ge, Song, Dang, Wang, Wang, Tang, et~al.]{bai2025qwen2}
Shuai Bai, Keqin Chen, Xuejing Liu, Jialin Wang, Wenbin Ge, Sibo Song, Kai Dang, Peng Wang, Shijie Wang, Jun Tang, et~al.
\newblock Qwen2. 5-vl technical report.
\newblock \emph{arXiv preprint arXiv:2502.13923}, 2025.

\bibitem[Bansal et~al.(2024)Bansal, Lin, Xie, Zong, Yarom, Bitton, Jiang, Sun, Chang, and Grover]{bansal2024videophy}
Hritik Bansal, Zongyu Lin, Tianyi Xie, Zeshun Zong, Michal Yarom, Yonatan Bitton, Chenfanfu Jiang, Yizhou Sun, Kai-Wei Chang, and Aditya Grover.
\newblock Videophy: Evaluating physical commonsense for video generation.
\newblock \emph{arXiv preprint arXiv:2406.03520}, 2024.

\bibitem[Bi et~al.(2024)Bi, Chen, Chen, Chen, Dai, Deng, Ding, Dong, Du, Fu, et~al.]{bi2024deepseek}
Xiao Bi, Deli Chen, Guanting Chen, Shanhuang Chen, Damai Dai, Chengqi Deng, Honghui Ding, Kai Dong, Qiushi Du, Zhe Fu, et~al.
\newblock Deepseek {LLM}: Scaling open-source language models with longtermism.
\newblock \emph{arXiv preprint arXiv:2401.02954}, 2024.

\bibitem[Blattmann et~al.(2023)Blattmann, Rombach, Ling, Dockhorn, Kim, Fidler, and Kreis]{blattmann2023align}
Andreas Blattmann, Robin Rombach, Huan Ling, Tim Dockhorn, Seung~Wook Kim, Sanja Fidler, and Karsten Kreis.
\newblock Align your latents: High-resolution video synthesis with latent diffusion models.
\newblock In \emph{CVPR}, 2023.

\bibitem[{ByteDance Seed}(2025)]{seed16}
{ByteDance Seed}.
\newblock Introducing to seed1.6 series, 2025.

\bibitem[Chen et~al.(2025{\natexlab{a}})Chen, Wu, Liu, Pan, Liu, Xie, Yu, and Ruan]{chen2025janus}
Xiaokang Chen, Zhiyu Wu, Xingchao Liu, Zizheng Pan, Wen Liu, Zhenda Xie, Xingkai Yu, and Chong Ruan.
\newblock Janus-pro: Unified multimodal understanding and generation with data and model scaling.
\newblock \emph{arXiv preprint arXiv:2501.17811}, 2025{\natexlab{a}}.

\bibitem[Chen et~al.(2025{\natexlab{b}})Chen, Wu, Liu, Pan, Liu, Xie, Yu, and Ruan]{janus}
Xiaokang Chen, Zhiyu Wu, Xingchao Liu, Zizheng Pan, Wen Liu, Zhenda Xie, Xingkai Yu, and Chong Ruan.
\newblock Janus-pro: Unified multimodal understanding and generation with data and model scaling.
\newblock \emph{arXiv preprint arXiv:2501.17811}, 2025{\natexlab{b}}.

\bibitem[Chen et~al.(2025{\natexlab{c}})Chen, Cao, Kag, Goel, Korolev, Jiang, Tulyakov, and Ren]{chen2025towards}
Yunuo Chen, Junli Cao, Anil Kag, Vidit Goel, Sergei Korolev, Chenfanfu Jiang, Sergey Tulyakov, and Jian Ren.
\newblock Towards physical understanding in video generation: A 3d point regularization approach.
\newblock \emph{arXiv preprint arXiv:2502.03639}, 2025{\natexlab{c}}.

\bibitem[Chen et~al.(2025{\natexlab{d}})Chen, Chen, Zhang, Huang, and Xie]{chen2025editboard}
Yupeng Chen, Penglin Chen, Xiaoyu Zhang, Yixian Huang, and Qian Xie.
\newblock Editboard: Towards a comprehensive evaluation benchmark for text-based video editing models.
\newblock In \emph{AAAI}, 2025{\natexlab{d}}.

\bibitem[Chen et~al.(2025{\natexlab{e}})Chen, Zhang, Hu, Zeng, Xue, He, Wang, Liu, Hu, and Yan]{chen2025ivebench}
Yinan Chen, Jiangning Zhang, Teng Hu, Yuxiang Zeng, Zhucun Xue, Qingdong He, Chengjie Wang, Yong Liu, Xiaobin Hu, and Shuicheng Yan.
\newblock Ivebench: Modern benchmark suite for instruction-guided video editing assessment.
\newblock \emph{arXiv preprint arXiv:2510.11647}, 2025{\natexlab{e}}.

\bibitem[Comanici et~al.(2025)Comanici, Bieber, Schaekermann, Pasupat, Sachdeva, Dhillon, Blistein, Ram, Zhang, Rosen, et~al.]{comanici2025gemini}
Gheorghe Comanici, Eric Bieber, Mike Schaekermann, Ice Pasupat, Noveen Sachdeva, Inderjit Dhillon, Marcel Blistein, Ori Ram, Dan Zhang, Evan Rosen, et~al.
\newblock Gemini 2.5: Pushing the frontier with advanced reasoning, multimodality, long context, and next generation agentic capabilities.
\newblock \emph{arXiv preprint arXiv:2507.06261}, 2025.

\bibitem[Deng et~al.(2025)Deng, Zhu, Li, Gou, Li, Wang, Zhong, Yu, Nie, Song, et~al.]{deng2025emerging}
Chaorui Deng, Deyao Zhu, Kunchang Li, Chenhui Gou, Feng Li, Zeyu Wang, Shu Zhong, Weihao Yu, Xiaonan Nie, Ziang Song, et~al.
\newblock Emerging properties in unified multimodal pretraining.
\newblock \emph{arXiv preprint arXiv:2505.14683}, 2025.

\bibitem[Duan et~al.(2025)Duan, Yu, Chen, Fei-Fei, and Wu]{duan2025worldscore}
Haoyi Duan, Hong-Xing Yu, Sirui Chen, Li Fei-Fei, and Jiajun Wu.
\newblock Worldscore: A unified evaluation benchmark for world generation.
\newblock \emph{arXiv preprint arXiv:2504.00983}, 2025.

\bibitem[Esser et~al.(2024)Esser, Kulal, Blattmann, Entezari, M{\"u}ller, Saini, Levi, Lorenz, Sauer, Boesel, et~al.]{esser2024scaling}
Patrick Esser, Sumith Kulal, Andreas Blattmann, Rahim Entezari, Jonas M{\"u}ller, Harry Saini, Yam Levi, Dominik Lorenz, Axel Sauer, Frederic Boesel, et~al.
\newblock Scaling rectified flow transformers for high-resolution image synthesis.
\newblock In \emph{ICML}, 2024.

\bibitem[Fan et~al.(2023)Fan, Luo, Gao, and Zhan]{fan2023aigcbench}
Fanda Fan, Chunjie Luo, Wanling Gao, and Jianfeng Zhan.
\newblock Aigcbench: Comprehensive evaluation of image-to-video content generated by ai.
\newblock \emph{BenchCouncil Transactions on Benchmarks, Standards and Evaluations}, 2023.

\bibitem[Feng et~al.(2024)Feng, Li, Saxon, Fu, Chen, and Wang]{feng2024tc}
Weixi Feng, Jiachen Li, Michael Saxon, Tsu-jui Fu, Wenhu Chen, and William~Yang Wang.
\newblock Tc-bench: Benchmarking temporal compositionality in text-to-video and image-to-video generation.
\newblock \emph{arXiv preprint arXiv:2406.08656}, 2024.

\bibitem[Fortin et~al.(2025)Fortin, Vernade, Kampf, and Reshi]{nano-banana-2025}
Alisa Fortin, Guillaume Vernade, Kat Kampf, and Ammaar Reshi.
\newblock Introducing gemini 2.5 flash image: our state-of-the-art image model, 2025.
\newblock Google Developers Blog, May 2025.

\bibitem[Gao et~al.(2024{\natexlab{a}})Gao, Zhang, Xu, Cai, and Shao]{gao2024flip}
Chongkai Gao, Haozhuo Zhang, Zhixuan Xu, Zhehao Cai, and Lin Shao.
\newblock Flip: Flow-centric generative planning as general-purpose manipulation world model.
\newblock \emph{arXiv preprint arXiv:2412.08261}, 2024{\natexlab{a}}.

\bibitem[Gao et~al.(2024{\natexlab{b}})Gao, Zhuo, Liu, Du, Luo, Qiu, Zhang, Lin, Huang, Geng, et~al.]{gao2024lumina}
Peng Gao, Le Zhuo, Dongyang Liu, Ruoyi Du, Xu Luo, Longtian Qiu, Yuhang Zhang, Chen Lin, Rongjie Huang, Shijie Geng, et~al.
\newblock Lumina-t2x: Transforming text into any modality, resolution, and duration via flow-based large diffusion transformers.
\newblock \emph{arXiv preprint arXiv:2405.05945}, 2024{\natexlab{b}}.

\bibitem[Gao et~al.(2025)Gao, Guo, Hoang, Huang, Jiang, Kong, Li, Li, Li, Li, et~al.]{gao2025seedance}
Yu Gao, Haoyuan Guo, Tuyen Hoang, Weilin Huang, Lu Jiang, Fangyuan Kong, Huixia Li, Jiashi Li, Liang Li, Xiaojie Li, et~al.
\newblock Seedance 1.0: Exploring the boundaries of video generation models.
\newblock \emph{arXiv preprint arXiv:2506.09113}, 2025.

\bibitem[GLM et~al.(2024)GLM, Zeng, Xu, Wang, Zhang, Yin, Zhang, Rojas, Feng, Zhao, et~al.]{glm2024chatglm}
Team GLM, Aohan Zeng, Bin Xu, Bowen Wang, Chenhui Zhang, Da Yin, Dan Zhang, Diego Rojas, Guanyu Feng, Hanlin Zhao, et~al.
\newblock {ChatGLM}: A family of large language models from glm-130b to glm-4 all tools.
\newblock \emph{arXiv preprint arXiv:2406.12793}, 2024.

\bibitem[{Google DeepMind}(2025)]{GoogleDeepMind2025Veo3}
{Google DeepMind}.
\newblock Veo-3 technical report.
\newblock Technical report, {Google DeepMind}, 2025.

\bibitem[Guo et~al.(2025{\natexlab{a}})Guo, Yang, Zhang, Song, Wang, Zhu, Xu, Zhang, Ma, Bi, et~al.]{guo2025deepseek}
Daya Guo, Dejian Yang, Haowei Zhang, Junxiao Song, Peiyi Wang, Qihao Zhu, Runxin Xu, Ruoyu Zhang, Shirong Ma, Xiao Bi, et~al.
\newblock Deepseek-r1 incentivizes reasoning in llms through reinforcement learning.
\newblock \emph{Nature}, 2025{\natexlab{a}}.

\bibitem[Guo et~al.(2023)Guo, Yang, Rao, Liang, Wang, Qiao, Agrawala, Lin, and Dai]{guo2023animatediff}
Yuwei Guo, Ceyuan Yang, Anyi Rao, Zhengyang Liang, Yaohui Wang, Yu Qiao, Maneesh Agrawala, Dahua Lin, and Bo Dai.
\newblock Animatediff: Animate your personalized text-to-image diffusion models without specific tuning.
\newblock \emph{arXiv preprint arXiv:2307.04725}, 2023.

\bibitem[Guo et~al.(2025{\natexlab{b}})Guo, Chen, Zhang, An, Qi, Jiang, Li, Zhang, Li, and Heng]{guo2025video}
Ziyu Guo, Xinyan Chen, Renrui Zhang, Ruichuan An, Yu Qi, Dongzhi Jiang, Xiangtai Li, Manyuan Zhang, Hongsheng Li, and Pheng-Ann Heng.
\newblock Are video models ready as zero-shot reasoners? an empirical study with the mme-cof benchmark.
\newblock \emph{arXiv preprint arXiv:2510.26802}, 2025{\natexlab{b}}.

\bibitem[Gupta et~al.(2024)Gupta, Yu, Sohn, Gu, Hahn, Li, Essa, Jiang, and Lezama]{gupta2024photorealistic}
Agrim Gupta, Lijun Yu, Kihyuk Sohn, Xiuye Gu, Meera Hahn, Fei-Fei Li, Irfan Essa, Lu Jiang, and Jos{\'e} Lezama.
\newblock Photorealistic video generation with diffusion models.
\newblock In \emph{ECCV}, 2024.

\bibitem[Henschel et~al.(2025)Henschel, Khachatryan, Poghosyan, Hayrapetyan, Tadevosyan, Wang, Navasardyan, and Shi]{henschel2025streamingt2v}
Roberto Henschel, Levon Khachatryan, Hayk Poghosyan, Daniil Hayrapetyan, Vahram Tadevosyan, Zhangyang Wang, Shant Navasardyan, and Humphrey Shi.
\newblock Streamingt2v: Consistent, dynamic, and extendable long video generation from text.
\newblock In \emph{CVPR}, 2025.

\bibitem[Ho et~al.(2020)Ho, Jain, and Abbeel]{ho2020denoising}
Jonathan Ho, Ajay Jain, and Pieter Abbeel.
\newblock Denoising diffusion probabilistic models.
\newblock In \emph{NeurIPS}, 2020.

\bibitem[Ho et~al.(2022{\natexlab{a}})Ho, Chan, Saharia, Whang, Gao, Gritsenko, Kingma, Poole, Norouzi, Fleet, et~al.]{ho2022imagen}
Jonathan Ho, William Chan, Chitwan Saharia, Jay Whang, Ruiqi Gao, Alexey Gritsenko, Diederik~P Kingma, Ben Poole, Mohammad Norouzi, David~J Fleet, et~al.
\newblock Imagen video: High definition video generation with diffusion models.
\newblock \emph{arXiv preprint arXiv:2210.02303}, 2022{\natexlab{a}}.

\bibitem[Ho et~al.(2022{\natexlab{b}})Ho, Salimans, Gritsenko, Chan, Norouzi, and Fleet]{ho2022video}
Jonathan Ho, Tim Salimans, Alexey Gritsenko, William Chan, Mohammad Norouzi, and David~J Fleet.
\newblock Video diffusion models.
\newblock In \emph{NeurIPS}, 2022{\natexlab{b}}.

\bibitem[Hong et~al.(2022)Hong, Ding, Zheng, Liu, and Tang]{hong2022cogvideo}
Wenyi Hong, Ming Ding, Wendi Zheng, Xinghan Liu, and Jie Tang.
\newblock Cogvideo: Large-scale pretraining for text-to-video generation via transformers.
\newblock \emph{arXiv preprint arXiv:2205.15868}, 2022.

\bibitem[Huang et~al.(2024)Huang, He, Yu, Zhang, Si, Jiang, Zhang, Wu, Jin, Chanpaisit, et~al.]{huang2024vbench}
Ziqi Huang, Yinan He, Jiashuo Yu, Fan Zhang, Chenyang Si, Yuming Jiang, Yuanhan Zhang, Tianxing Wu, Qingyang Jin, Nattapol Chanpaisit, et~al.
\newblock Vbench: Comprehensive benchmark suite for video generative models.
\newblock In \emph{CVPR}, 2024.

\bibitem[Huang et~al.(2025)Huang, Yu, Chen, Qiu, Debevec, and Liu]{huang2025vchain}
Ziqi Huang, Ning Yu, Gordon Chen, Haonan Qiu, Paul Debevec, and Ziwei Liu.
\newblock Vchain: Chain-of-visual-thought for reasoning in video generation.
\newblock \emph{arXiv preprint arXiv:2510.05094}, 2025.

\bibitem[Hurst et~al.(2024)Hurst, Lerer, Goucher, Perelman, Ramesh, Clark, Ostrow, Welihinda, Hayes, Radford, et~al.]{hurst2024gpt}
Aaron Hurst, Adam Lerer, Adam~P Goucher, Adam Perelman, Aditya Ramesh, Aidan Clark, AJ Ostrow, Akila Welihinda, Alan Hayes, Alec Radford, et~al.
\newblock Gpt-4o system card.
\newblock \emph{arXiv preprint arXiv:2410.21276}, 2024.

\bibitem[Jiang et~al.(2024)Jiang, Wu, Yang, Si, Lin, Qiao, Loy, and Liu]{jiang2024videobooth}
Yuming Jiang, Tianxing Wu, Shuai Yang, Chenyang Si, Dahua Lin, Yu Qiao, Chen~Change Loy, and Ziwei Liu.
\newblock Videobooth: Diffusion-based video generation with image prompts.
\newblock In \emph{CVPR}, 2024.

\bibitem[Jin et~al.(2024)Jin, Sun, Li, Xu, Jiang, Zhuang, Huang, Song, Mu, and Lin]{jin2024pyramidal}
Yang Jin, Zhicheng Sun, Ningyuan Li, Kun Xu, Hao Jiang, Nan Zhuang, Quzhe Huang, Yang Song, Yadong Mu, and Zhouchen Lin.
\newblock Pyramidal flow matching for efficient video generative modeling.
\newblock \emph{arXiv preprint arXiv:2410.05954}, 2024.

\bibitem[Kong et~al.(2024)Kong, Tian, Zhang, Min, Dai, Zhou, Xiong, Li, Wu, Zhang, et~al.]{kong2024hunyuanvideo}
Weijie Kong, Qi Tian, Zijian Zhang, Rox Min, Zuozhuo Dai, Jin Zhou, Jiangfeng Xiong, Xin Li, Bo Wu, Jianwei Zhang, et~al.
\newblock Hunyuanvideo: A systematic framework for large video generative models.
\newblock \emph{arXiv preprint arXiv:2412.03603}, 2024.

\bibitem[Labs et~al.(2025)Labs, Batifol, Blattmann, Boesel, Consul, Diagne, Dockhorn, English, English, Esser, et~al.]{labs2025flux}
Black~Forest Labs, Stephen Batifol, Andreas Blattmann, Frederic Boesel, Saksham Consul, Cyril Diagne, Tim Dockhorn, Jack English, Zion English, Patrick Esser, et~al.
\newblock Flux. 1 kontext: Flow matching for in-context image generation and editing in latent space.
\newblock \emph{arXiv preprint arXiv:2506.15742}, 2025.

\bibitem[Li et~al.(2025{\natexlab{a}})Li, Jiang, Xiao, Wang, Yi, Wu, and Cai]{li2025magicid}
Hengjia Li, Lifan Jiang, Xi Xiao, Tianyang Wang, Hongwei Yi, Boxi Wu, and Deng Cai.
\newblock Magicid: Hybrid preference optimization for id-consistent and dynamic-preserved video customization.
\newblock \emph{arXiv preprint arXiv:2503.12689}, 2025{\natexlab{a}}.

\bibitem[Li et~al.(2025{\natexlab{b}})Li, Qiu, Zhang, Wang, Wei, Li, Zhang, Wu, and Cai]{li2025personalvideo}
Hengjia Li, Haonan Qiu, Shiwei Zhang, Xiang Wang, Yujie Wei, Zekun Li, Yingya Zhang, Boxi Wu, and Deng Cai.
\newblock Personalvideo: High id-fidelity video customization without dynamic and semantic degradation.
\newblock In \emph{ICCV}, 2025{\natexlab{b}}.

\bibitem[Li et~al.(2025{\natexlab{c}})Li, Wang, Hu, Huang, Chen, Ou, Tao, Wan, Qi, and Feng]{li2025easier}
Ouxiang Li, Yuan Wang, Xinting Hu, Huijuan Huang, Rui Chen, Jiarong Ou, Xin Tao, Pengfei Wan, Xiaojuan Qi, and Fuli Feng.
\newblock Easier painting than thinking: Can text-to-image models set the stage, but not direct the play?
\newblock \emph{arXiv preprint arXiv:2509.03516}, 2025{\natexlab{c}}.

\bibitem[Li et~al.(2024)Li, Qiu, Chen, Kuen, Lin, Singh, and Raj]{li2024controlvar}
Xiang Li, Kai Qiu, Hao Chen, Jason Kuen, Zhe Lin, Rita Singh, and Bhiksha Raj.
\newblock Controlvar: Exploring controllable visual autoregressive modeling.
\newblock \emph{arXiv preprint arXiv:2406.09750}, 2024.

\bibitem[Lin et~al.(2024)Lin, Ge, Cheng, Li, Zhu, Wang, He, Ye, Yuan, Chen, et~al.]{lin2024open}
Bin Lin, Yunyang Ge, Xinhua Cheng, Zongjian Li, Bin Zhu, Shaodong Wang, Xianyi He, Yang Ye, Shenghai Yuan, Liuhan Chen, et~al.
\newblock Open-sora plan: Open-source large video generation model.
\newblock \emph{arXiv preprint arXiv:2412.00131}, 2024.

\bibitem[Liu et~al.(2025)Liu, Liu, Liang, Yuan, Liu, Zheng, Wu, Wang, Xia, Wang, et~al.]{liu2025improving}
Jie Liu, Gongye Liu, Jiajun Liang, Ziyang Yuan, Xiaokun Liu, Mingwu Zheng, Xiele Wu, Qiulin Wang, Menghan Xia, Xintao Wang, et~al.
\newblock Improving video generation with human feedback.
\newblock \emph{arXiv preprint arXiv:2501.13918}, 2025.

\bibitem[Liu et~al.(2024{\natexlab{a}})Liu, Ren, Gupta, and Wang]{liu2024physgen}
Shaowei Liu, Zhongzheng Ren, Saurabh Gupta, and Shenlong Wang.
\newblock Physgen: Rigid-body physics-grounded image-to-video generation.
\newblock In \emph{ECCV}, 2024{\natexlab{a}}.

\bibitem[Liu et~al.(2023)Liu, Li, Ren, Gao, Li, Chen, Sun, and Hou]{liu2023fetv}
Yuanxin Liu, Lei Li, Shuhuai Ren, Rundong Gao, Shicheng Li, Sishuo Chen, Xu Sun, and Lu Hou.
\newblock Fetv: A benchmark for fine-grained evaluation of open-domain text-to-video generation.
\newblock \emph{NeurIPS}, 2023.

\bibitem[Liu et~al.(2024{\natexlab{b}})Liu, Cun, Liu, Wang, Zhang, Chen, Liu, Zeng, Chan, and Shan]{liu2024evalcrafter}
Yaofang Liu, Xiaodong Cun, Xuebo Liu, Xintao Wang, Yong Zhang, Haoxin Chen, Yang Liu, Tieyong Zeng, Raymond Chan, and Ying Shan.
\newblock Evalcrafter: Benchmarking and evaluating large video generation models.
\newblock In \emph{CVPR}, 2024{\natexlab{b}}.

\bibitem[Ma et~al.(2024)Ma, Wang, Jia, Chen, Liu, Li, Chen, and Qiao]{ma2024latte}
Xin Ma, Yaohui Wang, Gengyun Jia, Xinyuan Chen, Ziwei Liu, Yuan-Fang Li, Cunjian Chen, and Yu Qiao.
\newblock Latte: Latent diffusion transformer for video generation.
\newblock \emph{arXiv preprint arXiv:2401.03048}, 2024.

\bibitem[Meng et~al.(2024)Meng, Liao, Tan, Shao, Lu, Zhang, Cheng, Li, Qiao, and Luo]{meng2024towards}
Fanqing Meng, Jiaqi Liao, Xinyu Tan, Wenqi Shao, Quanfeng Lu, Kaipeng Zhang, Yu Cheng, Dianqi Li, Yu Qiao, and Ping Luo.
\newblock Towards world simulator: Crafting physical commonsense-based benchmark for video generation.
\newblock \emph{arXiv preprint arXiv:2410.05363}, 2024.

\bibitem[Montanaro et~al.(2024)Montanaro, Savant~Aira, Aiello, Valsesia, and Magli]{montanaro2024motioncraft}
Antonio Montanaro, Luca Savant~Aira, Emanuele Aiello, Diego Valsesia, and Enrico Magli.
\newblock Motioncraft: Physics-based zero-shot video generation.
\newblock In \emph{NeurIPS}, 2024.

\bibitem[Nichol and Dhariwal(2021)]{nichol2021improved}
Alexander~Quinn Nichol and Prafulla Dhariwal.
\newblock Improved denoising diffusion probabilistic models.
\newblock In \emph{ICML}, 2021.

\bibitem[{OpenAI}(2025{\natexlab{a}})]{openai2025gpt5}
{OpenAI}.
\newblock Gpt-5 system card.
\newblock Technical report, OpenAI, 2025{\natexlab{a}}.

\bibitem[{OpenAI}(2025{\natexlab{b}})]{openai2025o3}
{OpenAI}.
\newblock Openai o3 and o4-mini system card.
\newblock Technical report, OpenAI, 2025{\natexlab{b}}.

\bibitem[{OpenAI}(2025{\natexlab{c}})]{openai2025sora2}
{OpenAI}.
\newblock Sora 2 system card.
\newblock Technical report, OpenAI, 2025{\natexlab{c}}.

\bibitem[Peebles and Xie(2023)]{peebles2023scalable}
William Peebles and Saining Xie.
\newblock Scalable diffusion models with transformers.
\newblock In \emph{ICCV}, 2023.

\bibitem[{PixVerse Team}(2025)]{pixverse-v5-2025}
{PixVerse Team}.
\newblock Pixverse v5, 2025.

\bibitem[Rombach et~al.(2022)Rombach, Blattmann, Lorenz, Esser, and Ommer]{rombach2022high}
Robin Rombach, Andreas Blattmann, Dominik Lorenz, Patrick Esser, and Bj{\"o}rn Ommer.
\newblock High-resolution image synthesis with latent diffusion models.
\newblock In \emph{CVPR}, 2022.

\bibitem[Song et~al.(2020{\natexlab{a}})Song, Meng, and Ermon]{song2020denoising}
Jiaming Song, Chenlin Meng, and Stefano Ermon.
\newblock Denoising diffusion implicit models.
\newblock \emph{arXiv preprint arXiv:2010.02502}, 2020{\natexlab{a}}.

\bibitem[Song et~al.(2020{\natexlab{b}})Song, Sohl-Dickstein, Kingma, Kumar, Ermon, and Poole]{song2020score}
Yang Song, Jascha Sohl-Dickstein, Diederik~P Kingma, Abhishek Kumar, Stefano Ermon, and Ben Poole.
\newblock Score-based generative modeling through stochastic differential equations.
\newblock \emph{arXiv preprint arXiv:2011.13456}, 2020{\natexlab{b}}.

\bibitem[Sun et~al.(2025{\natexlab{a}})Sun, Huang, Liu, Wu, Xu, Li, and Liu]{sun2025t2v}
Kaiyue Sun, Kaiyi Huang, Xian Liu, Yue Wu, Zihan Xu, Zhenguo Li, and Xihui Liu.
\newblock T2v-compbench: A comprehensive benchmark for compositional text-to-video generation.
\newblock In \emph{CVPR}, 2025{\natexlab{a}}.

\bibitem[Sun et~al.(2025{\natexlab{b}})Sun, Liang, Fan, Gao, and Gao]{sun2025ve}
Shangkun Sun, Xiaoyu Liang, Songlin Fan, Wenxu Gao, and Wei Gao.
\newblock Ve-bench: Subjective-aligned benchmark suite for text-driven video editing quality assessment.
\newblock In \emph{AAAI}, 2025{\natexlab{b}}.

\bibitem[Teng et~al.(2025)Teng, Jia, Sun, Li, Li, Tang, Han, Zhang, Zhang, Luo, et~al.]{teng2025magi}
Hansi Teng, Hongyu Jia, Lei Sun, Lingzhi Li, Maolin Li, Mingqiu Tang, Shuai Han, Tianning Zhang, WQ Zhang, Weifeng Luo, et~al.
\newblock Magi-1: Autoregressive video generation at scale.
\newblock \emph{arXiv preprint arXiv:2505.13211}, 2025.

\bibitem[Tian et~al.(2024)Tian, Jiang, Yuan, Peng, and Wang]{tian2024visual}
Keyu Tian, Yi Jiang, Zehuan Yuan, Bingyue Peng, and Liwei Wang.
\newblock Visual autoregressive modeling: Scalable image generation via next-scale prediction.
\newblock In \emph{NeurIPS}, 2024.

\bibitem[Touvron et~al.(2023)Touvron, Lavril, Izacard, Martinet, Lachaux, Lacroix, Rozi{\`e}re, Goyal, Hambro, Azhar, et~al.]{touvron2023llama}
Hugo Touvron, Thibaut Lavril, Gautier Izacard, Xavier Martinet, Marie-Anne Lachaux, Timoth{\'e}e Lacroix, Baptiste Rozi{\`e}re, Naman Goyal, Eric Hambro, Faisal Azhar, et~al.
\newblock {LLaMA}: Open and efficient foundation language models.
\newblock \emph{arXiv preprint arXiv:2302.13971}, 2023.

\bibitem[van~den Oord et~al.(2024)van~den Oord, Roman, Lacombe, Fortin, and Vernade]{google-veo2-2025}
A{\"a}ron van~den Oord, Elias Roman, Olivier Lacombe, Alisa Fortin, and Guillaume Vernade.
\newblock Veo 2, 2024.
\newblock Google Research Blog, December 2024.

\bibitem[von Platen et~al.(2022)von Platen, Patil, Lozhkov, Cuenca, Lambert, Rasul, Davaadorj, Nair, Paul, Berman, Xu, Liu, and Wolf]{von-platen-etal-2022-diffusers}
Patrick von Platen, Suraj Patil, Anton Lozhkov, Pedro Cuenca, Nathan Lambert, Kashif Rasul, Mishig Davaadorj, Dhruv Nair, Sayak Paul, William Berman, Yiyi Xu, Steven Liu, and Thomas Wolf.
\newblock Diffusers: State-of-the-art diffusion models, 2022.

\bibitem[Wan et~al.(2025)Wan, Wang, Ai, Wen, Mao, Xie, Chen, Yu, Zhao, Yang, Zeng, Wang, Zhang, Zhou, Wang, Chen, Zhu, Zhao, Yan, Huang, Feng, Zhang, Li, Wu, Chu, Feng, Zhang, Sun, Fang, Wang, Gui, Weng, Shen, Lin, Wang, Wang, Zhou, Wang, Shen, Yu, Shi, Huang, Xu, Kou, Lv, Li, Liu, Wang, Zhang, Huang, Li, Wu, Liu, Pan, Zheng, Hong, Shi, Feng, Jiang, Han, Wu, and Liu]{wan2025}
Team Wan, Ang Wang, Baole Ai, Bin Wen, Chaojie Mao, Chen-Wei Xie, Di Chen, Feiwu Yu, Haiming Zhao, Jianxiao Yang, Jianyuan Zeng, Jiayu Wang, Jingfeng Zhang, Jingren Zhou, Jinkai Wang, Jixuan Chen, Kai Zhu, Kang Zhao, Keyu Yan, Lianghua Huang, Mengyang Feng, Ningyi Zhang, Pandeng Li, Pingyu Wu, Ruihang Chu, Ruili Feng, Shiwei Zhang, Siyang Sun, Tao Fang, Tianxing Wang, Tianyi Gui, Tingyu Weng, Tong Shen, Wei Lin, Wei Wang, Wei Wang, Wenmeng Zhou, Wente Wang, Wenting Shen, Wenyuan Yu, Xianzhong Shi, Xiaoming Huang, Xin Xu, Yan Kou, Yangyu Lv, Yifei Li, Yijing Liu, Yiming Wang, Yingya Zhang, Yitong Huang, Yong Li, You Wu, Yu Liu, Yulin Pan, Yun Zheng, Yuntao Hong, Yupeng Shi, Yutong Feng, Zeyinzi Jiang, Zhen Han, Zhi-Fan Wu, and Ziyu Liu.
\newblock Wan: Open and advanced large-scale video generative models.
\newblock \emph{arXiv preprint arXiv:2503.20314}, 2025.

\bibitem[Wang et~al.(2025{\natexlab{a}})Wang, Ma, Cao, Zheng, Zhang, Feng, Liu, Ma, Cheng, Leng, et~al.]{wang2025wisa}
Jing Wang, Ao Ma, Ke Cao, Jun Zheng, Zhanjie Zhang, Jiasong Feng, Shanyuan Liu, Yuhang Ma, Bo Cheng, Dawei Leng, et~al.
\newblock Wisa: World simulator assistant for physics-aware text-to-video generation.
\newblock \emph{arXiv preprint arXiv:2503.08153}, 2025{\natexlab{a}}.

\bibitem[Wang et~al.(2025{\natexlab{b}})Wang, Chen, Ma, Zhou, Huang, Wang, Yang, He, Yu, Yang, et~al.]{wang2025lavie}
Yaohui Wang, Xinyuan Chen, Xin Ma, Shangchen Zhou, Ziqi Huang, Yi Wang, Ceyuan Yang, Yinan He, Jiashuo Yu, Peiqing Yang, et~al.
\newblock Lavie: High-quality video generation with cascaded latent diffusion models.
\newblock \emph{IJCV}, 2025{\natexlab{b}}.

\bibitem[Wang et~al.(2025{\natexlab{c}})Wang, Li, Yan, He, Yu, Zeng, Wang, Ma, Huang, Gao, et~al.]{wang2025internvideo2}
Yi Wang, Xinhao Li, Ziang Yan, Yinan He, Jiashuo Yu, Xiangyu Zeng, Chenting Wang, Changlian Ma, Haian Huang, Jianfei Gao, et~al.
\newblock Internvideo2. 5: Empowering video mllms with long and rich context modeling.
\newblock \emph{arXiv preprint arXiv:2501.12386}, 2025{\natexlab{c}}.

\bibitem[Wei et~al.(2025)Wei, Zhang, Yuan, Gong, Tang, Wang, Qiu, Li, Tan, Zhang, et~al.]{wei2025dreamrelation}
Yujie Wei, Shiwei Zhang, Hangjie Yuan, Biao Gong, Longxiang Tang, Xiang Wang, Haonan Qiu, Hengjia Li, Shuai Tan, Yingya Zhang, et~al.
\newblock Dreamrelation: Relation-centric video customization.
\newblock \emph{arXiv preprint arXiv:2503.07602}, 2025.

\bibitem[Wiedemer et~al.(2025)Wiedemer, Li, Vicol, Gu, Matarese, Swersky, Kim, Jaini, and Geirhos]{wiedemer2025video}
Thadd{\"a}us Wiedemer, Yuxuan Li, Paul Vicol, Shixiang~Shane Gu, Nick Matarese, Kevin Swersky, Been Kim, Priyank Jaini, and Robert Geirhos.
\newblock Video models are zero-shot learners and reasoners.
\newblock \emph{arXiv preprint arXiv:2509.20328}, 2025.

\bibitem[Wu et~al.(2025{\natexlab{a}})Wu, Zhang, Wang, Zhou, Zheng, Qi, Shan, and Li]{wu2025customcrafter}
Tao Wu, Yong Zhang, Xintao Wang, Xianpan Zhou, Guangcong Zheng, Zhongang Qi, Ying Shan, and Xi Li.
\newblock Customcrafter: Customized video generation with preserving motion and concept composition abilities.
\newblock In \emph{AAAI}, 2025{\natexlab{a}}.

\bibitem[Wu et~al.(2025{\natexlab{b}})Wu, Li, Hu, Ye, Zeng, Yu, Zhu, Schiele, Yang, and Yang]{wu2025kris}
Yongliang Wu, Zonghui Li, Xinting Hu, Xinyu Ye, Xianfang Zeng, Gang Yu, Wenbo Zhu, Bernt Schiele, Ming-Hsuan Yang, and Xu Yang.
\newblock Kris-bench: Benchmarking next-level intelligent image editing models.
\newblock \emph{arXiv preprint arXiv:2505.16707}, 2025{\natexlab{b}}.

\bibitem[{xAI}(2025)]{grok4}
{xAI}.
\newblock Grok 4 model card.
\newblock Technical report, xAI, 2025.

\bibitem[Xie et~al.(2024)Xie, Mao, Bai, Zhang, Wang, Lin, Gu, Chen, Yang, and Shou]{xie2024show}
Jinheng Xie, Weijia Mao, Zechen Bai, David~Junhao Zhang, Weihao Wang, Kevin~Qinghong Lin, Yuchao Gu, Zhijie Chen, Zhenheng Yang, and Mike~Zheng Shou.
\newblock Show-o: One single transformer to unify multimodal understanding and generation.
\newblock \emph{arXiv preprint arXiv:2408.12528}, 2024.

\bibitem[Xie et~al.(2025)Xie, Zhao, Jiang, and Jiang]{xie2025physanimator}
Tianyi Xie, Yiwei Zhao, Ying Jiang, and Chenfanfu Jiang.
\newblock Physanimator: Physics-guided generative cartoon animation.
\newblock In \emph{CVPR}, 2025.

\bibitem[Xue et~al.(2025)Xue, Yin, Yang, and Gao]{xue2025phyt2v}
Qiyao Xue, Xiangyu Yin, Boyuan Yang, and Wei Gao.
\newblock Phyt2v: Llm-guided iterative self-refinement for physics-grounded text-to-video generation.
\newblock In \emph{CVPR}, 2025.

\bibitem[Yang et~al.(2024{\natexlab{a}})Yang, Yang, Zhang, Hui, Zheng, Yu, Li, Liu, Huang, Wei, et~al.]{yang2024qwen2llm}
An Yang, Baosong Yang, Beichen Zhang, Binyuan Hui, Bo Zheng, Bowen Yu, Chengyuan Li, Dayiheng Liu, Fei Huang, Haoran Wei, et~al.
\newblock Qwen2.5 technical report.
\newblock \emph{arXiv preprint arXiv:2412.15115}, 2024{\natexlab{a}}.

\bibitem[Yang et~al.(2024{\natexlab{b}})Yang, Teng, Zheng, Ding, Huang, Xu, Yang, Hong, Zhang, Feng, et~al.]{yang2024cogvideox}
Zhuoyi Yang, Jiayan Teng, Wendi Zheng, Ming Ding, Shiyu Huang, Jiazheng Xu, Yuanming Yang, Wenyi Hong, Xiaohan Zhang, Guanyu Feng, et~al.
\newblock Cogvideox: Text-to-video diffusion models with an expert transformer.
\newblock \emph{arXiv preprint arXiv:2408.06072}, 2024{\natexlab{b}}.

\bibitem[Ye et~al.(2025)Ye, Huang, Wang, Wan, Zhang, and Luo]{ye2025stylemaster}
Zixuan Ye, Huijuan Huang, Xintao Wang, Pengfei Wan, Di Zhang, and Wenhan Luo.
\newblock Stylemaster: Stylize your video with artistic generation and translation.
\newblock In \emph{CVPR}, 2025.

\bibitem[Yuan et~al.(2025)Yuan, Wang, Wickremasinghe, Nadir, Ma, and Chan]{yuan2025newtongen}
Yu Yuan, Xijun Wang, Tharindu Wickremasinghe, Zeeshan Nadir, Bole Ma, and Stanley~H Chan.
\newblock Newtongen: Physics-consistent and controllable text-to-video generation via neural newtonian dynamics.
\newblock \emph{arXiv preprint arXiv:2509.21309}, 2025.

\bibitem[Zhang et~al.(2025{\natexlab{a}})Zhang, Lei, Kong, Wang, Xu, Song, Guo, Liu, Li, and Chen]{zhang2025ui2v}
Ailing Zhang, Lina Lei, Dehong Kong, Zhixin Wang, Jiaqi Xu, Fenglong Song, Chun-Le Guo, Chang Liu, Fan Li, and Jie Chen.
\newblock Ui2v-bench: An understanding-based image-to-video generation benchmark.
\newblock \emph{arXiv preprint arXiv:2509.24427}, 2025{\natexlab{a}}.

\bibitem[Zhang et~al.(2023)Zhang, Rao, and Agrawala]{zhang2023adding}
Lvmin Zhang, Anyi Rao, and Maneesh Agrawala.
\newblock Adding conditional control to text-to-image diffusion models.
\newblock In \emph{CVPR}, 2023.

\bibitem[Zhang et~al.(2024)Zhang, Yu, Wu, Feng, Zheng, Snavely, Wu, and Freeman]{zhang2024physdreamer}
Tianyuan Zhang, Hong-Xing Yu, Rundi Wu, Brandon~Y Feng, Changxi Zheng, Noah Snavely, Jiajun Wu, and William~T Freeman.
\newblock Physdreamer: Physics-based interaction with 3d objects via video generation.
\newblock In \emph{ECCV}, 2024.

\bibitem[Zhang et~al.(2025{\natexlab{b}})Zhang, Liao, Zhang, Meng, Wan, Yan, and Cheng]{zhang2025videorepa}
Xiangdong Zhang, Jiaqi Liao, Shaofeng Zhang, Fanqing Meng, Xiangpeng Wan, Junchi Yan, and Yu Cheng.
\newblock Videorepa: Learning physics for video generation through relational alignment with foundation models.
\newblock \emph{arXiv preprint arXiv:2505.23656}, 2025{\natexlab{b}}.

\bibitem[Zhang et~al.(2025{\natexlab{c}})Zhang, Guo, Pan, Yao, Zhu, Jiang, Wang, Jin, and Zhao]{zhang2025tcsinger}
Yu Zhang, Wenxiang Guo, Changhao Pan, Dongyu Yao, Zhiyuan Zhu, Ziyue Jiang, Yuhan Wang, Tao Jin, and Zhou Zhao.
\newblock Tcsinger 2: Customizable multilingual zero-shot singing voice synthesis.
\newblock \emph{arXiv preprint arXiv:2505.14910}, 2025{\natexlab{c}}.

\bibitem[Zhang et~al.(2025{\natexlab{d}})Zhang, Guo, Pan, Zhu, Li, Lu, Huang, Zhang, Hong, Jiang, et~al.]{zhang2025versatile}
Yu Zhang, Wenxiang Guo, Changhao Pan, Zhiyuan Zhu, Ruiqi Li, Jingyu Lu, Rongjie Huang, Ruiyuan Zhang, Zhiqing Hong, Ziyue Jiang, et~al.
\newblock Versatile framework for song generation with prompt-based control.
\newblock \emph{arXiv preprint arXiv:2504.19062}, 2025{\natexlab{d}}.

\bibitem[Zhao et~al.(2025)Zhao, Zhang, Tang, Zhu, Li, Chai, Zhang, Xia, Zhai, Yan, et~al.]{zhao2025envisioning}
Xiangyu Zhao, Peiyuan Zhang, Kexian Tang, Xiaorong Zhu, Hao Li, Wenhao Chai, Zicheng Zhang, Renqiu Xia, Guangtao Zhai, Junchi Yan, et~al.
\newblock Envisioning beyond the pixels: Benchmarking reasoning-informed visual editing.
\newblock \emph{arXiv preprint arXiv:2504.02826}, 2025.

\bibitem[Zheng et~al.(2025)Zheng, Huang, Liu, Zou, He, Zhang, Gu, Zhang, He, Zheng, et~al.]{zheng2025vbench}
Dian Zheng, Ziqi Huang, Hongbo Liu, Kai Zou, Yinan He, Fan Zhang, Lulu Gu, Yuanhan Zhang, Jingwen He, Wei-Shi Zheng, et~al.
\newblock Vbench-2.0: Advancing video generation benchmark suite for intrinsic faithfulness.
\newblock \emph{arXiv preprint arXiv:2503.21755}, 2025.

\bibitem[Zi et~al.(2025)Zi, Ruan, Chen, Qi, Hao, Zhao, Huang, Liang, Xiao, and Wong]{zi2025se}
Bojia Zi, Penghui Ruan, Marco Chen, Xianbiao Qi, Shaozhe Hao, Shihao Zhao, Youze Huang, Bin Liang, Rong Xiao, and Kam-Fai Wong.
\newblock Se$\backslash$\~{} norita-2m: A high-quality instruction-based dataset for general video editing by video specialists.
\newblock \emph{arXiv preprint arXiv:2502.06734}, 2025.

\end{thebibliography}
